\documentclass{article}
\usepackage[final]{neurips_2025}
\usepackage{natbib}
\usepackage[utf8]{inputenc}
\usepackage[T1]{fontenc}
\usepackage{tabularray}
\usepackage{amsmath}         % AMS math package
\usepackage{amsfonts}        % For \mathbb
\usepackage{amssymb}         % For other math 
\usepackage{bm}              % For bold math \bm{}
\usepackage{nicefrac}
\usepackage{fix-cm}

\usepackage{graphicx}
\usepackage[skip=0pt]{caption} % For customizing captions
\usepackage{subcaption}        % For subfigures
\usepackage{placeins} % 包含 placeins 宏包

\usepackage{booktabs}        % For professional tables
\usepackage{makecell}        % For multi-line cells in tables
\usepackage{multirow}        % For multi-row cells in tables
\usepackage{tabularx}        % For fixed-width tables
\usepackage{longtable}       % For tables spanning pages
\usepackage{array}           % For advanced column formatting
\usepackage{ragged2e}        % For improved ragged alignment
\usepackage[table,xcdraw]{xcolor} % For table coloring (loaded only once with options)

\usepackage{listings}        % For displaying code

\usepackage[linesnumbered,ruled,vlined]{algorithm2e}

\usepackage{hyperref}          % Should be loaded last generally
\usepackage{wrapfig}         % For text wrapping around figures
\usepackage{xspace}
\usepackage{microtype}

\newcommand{\mname}{{\sc lager}\xspace} 

\title{Beyond the Surface: Enhancing LLM-as-a-Judge Alignment with Human via Internal Representations}

\author{%
Peng Lai$^{1}$, \
Jianjie Zheng$^{1}$, \
Sijie Cheng$^{2}$,
Yun Chen$^{3}$,
Peng Li$^{2}$,
Yang Liu$^{2}$,
Guanhua Chen$^{1}\thanks{\ \ Corresponding author.}$\\
$^1$Southern University of Science and Technology, 
$^2$Tsinghua University \\
$^3$Shanghai University of Finance and Economics  \\
}

\begin{document}

\maketitle
\begin{abstract}
The growing scale of evaluation tasks has led to the widespread adoption of automated evaluation using LLMs, a paradigm known as "LLM-as-a-judge". However, improving its alignment with human preferences without complex prompts or fine-tuning remains challenging. Previous studies mainly optimize based on shallow outputs, overlooking rich cross-layer representations. In this work, motivated by preliminary findings that middle-to-upper layers encode semantically and task-relevant representations that are often more aligned with human judgments than the final layer, we propose \mname, a \textbf{post-hoc, plug-and-play} framework for improving the alignment of LLM-as-a-Judge point-wise evaluations with human scores, by leveraging internal representations. \mname produces fine-grained judgment scores by aggregating cross-layer score-token logits and computing the expected score from a softmax-based distribution, while keeping the LLM backbone frozen and ensuring no impact on the inference process.
\mname fully leverages the complementary information across different layers, overcoming the limitations of relying solely on the final layer.
% Experiments demonstrate the superior performance of the PalmScore method over baseline methods across Flask, HelpSteer, and BIGGen bench. 
We evaluate our method on the standard alignment benchmarks Flask, HelpSteer, and BIGGen using Spearman correlation, and find that \mname achieves improvements of up to 7.5\% over the best baseline across these benchmarks.
Without reasoning steps, \mname matches or outperforms reasoning-based methods. Experiments on downstream applications, such as data selection and emotional understanding, further show the generalization of \mname.
\end{abstract}

\section{Introduction} \label{sec:intro}

%% 介绍evaluation/judge的重要性，传统的方法比如classifier/BLEU score等，现在基于LLM的方法。
% Recently, large language models have been widely used for automated evaluations owing to their powerful capabilities~\citep{LLMs-instead-of-Human-Judges,eval-Toxicity,Evaluation-of-Free-Form-Text}. The large language model acts as a judge to score the quality of the responses generated by the LLMs. Specifically, a large language model judge will be prompted under the guidance of the assessment task. The LLM judge will be given a question and an answer to this question (or an instruction and a response to this instruction). The judge is asked to rate the answer (or response) according to some scoring criteria and other specific requirements. Compared with human evaluation, it reduces the need for human involvement and enables scalable benchmarks and rapid iterations~\citep{difference}. In addition, the LLM can provide not only scores but also explanations. Automated evaluations using large language models tend to replace human judgments~\citep{LLMs-instead-of-Human-Judges}.
% Conventional text evaluation metrics \citep{papineni-etal-2002-bleu,lin-2004-rouge,banerjee-lavie-2005-meteor} struggle to fully capture the semantic and contextual complexity especially in open-ended tasks~\citep{g-eval,LLM-eval}.
% While embedding-based similarity methods \citep{bertscore,bartscore,gptscore} can identify semantic similarities, they still fall short in assessing complex textual relationships~\citep{LLM-eval,10.1145/3625007.3627305}. 

%% llm的发展和优势，llm-judge的介绍
Large language models (LLMs) have witnessed significant progress in various tasks like math reasoning \citep{math1,duan2024multitoolintegrationapplicationmath} and open-domain question answering \citep{allemang2024increasingllmaccuracyquestion},and exhibit great generalization and reasoning abilities on unseen tasks and domains \citep{li-etal-2024-fundamental,ye-2024-cross,ni2024smalllanguagemodeldata}, which makes it possible to evaluate the model generations with an LLM \citep{LLMs-instead-of-Human-Judges,Judging-LLM-as-a-Judge}. 
LLM-as-a-judge \citep{LLM-eval,li2024llmsasjudgescomprehensivesurveyllmbased,wu2024metarewardinglanguagemodelsselfimproving,skyworkcritic2024}, an emerging approach to text evaluation, uses large language models to perform automated and scalable assessment of textual responses, reducing the reliance on human annotation.
% As an emerging paradigm for text evaluation, LLM-as-a-judge \citep{LLM-eval,li2024llmsasjudgescomprehensivesurveyllmbased,wu2024metarewardinglanguagemodelsselfimproving,skyworkcritic2024} provides more fine-grained and reliable evaluations that significantly enhance the effectiveness and applicability of automated assessments. %~\citep{song2024finesurefinegrainedsummarizationevaluation}. 
% llm-judge的应用场景. 
This paradigm has found widespread applications in various scenarios such as model evaluation \citep{LLM-eval}, data synthesis \citep{Prometheus,ultrafeedback,helpsteer2}, and model enhancement via verification and critic agents \citep{pace2024westofnsyntheticpreferencesselfimproving,Evaluation-of-Free-Form-Text}. 

% where no reference response is available. The model responses are evaluated from different aspects with an LLM judge to evaluate and compare the model performance \citep{LLM-eval}. The LLM judge is also used to create preference data \citep{wu2024thinkingllmsgeneralinstruction,Prometheus}, filter labeled data \citep{ultrafeedback,helpsteer2}, and provide the reward information \citep{ye2024conj,hu-etal-2024-themis} for finetuning. During inference, LLM judge can serve as the verifier \citep{pace2024westofnsyntheticpreferencesselfimproving} in Best-of-N sampling or provide feedback \citep{Evaluation-of-Free-Form-Text} to iteratively refine the responses. 

% 已有的研究工作
Researchers have explored various approaches to improve the consistency of judgment with human experts. Prompt-based methods \citep{LLM-eval,g-eval} improve the judge performance with guidelines encouraging step-by-step reasoning \citep{feedback-first,LLM-eval,zhuo2024icescoreinstructinglargelanguage} or carefully curated examples \citep{song2025manyshotincontextlearninghelp}. However, these methods rely on chain-of-thought reasoning, which introduces additional computational cost \citep{gu2024surveyllmasajudge}. Other researchers propose to finetune a LLM with a specialized dataset \citep{flask,cui2023ultrafeedback}to adapt the LLM to the judgment task. Although effective, these methods face the generalization problem \citep{doostmohammadi2024reliableautomaticevaluationmethods} as they have been adapted to the domain of the training data. 

%% 我们的工作，方法介绍，为什么效果更好，实验结果简述
In this paper, we propose an effective LLM-as-a-judge framework called \mname, to enhance \underline{\textbf{L}}LM-as-a-Judge \underline{\textbf{a}}li\underline{\textbf{g}}nment with human via int\underline{\textbf{e}}rnal \underline{\textbf{r}}epresentations. Inspired by prior studies \citep{wang-etal-2020-layer,yang2023gtrans,roh2024levi} showing that middle-to-upper layers encode richer semantic and task-specific information, we propose to aggregate the logits of score tokens from different layers. By applying a weighted combination followed by a softmax, we obtain a probability distribution over candidate scores, from which the final score is computed as the expected value.  \mname leverages the semantic diversity across different layers, allowing the final score to integrate both low-level lexical cues and high-level reasoning signals, thereby providing a more comprehensive reflection of the model's understanding of the scoring task.
\textbf{\mname is plug-and-play, leaving model parameters and next-token predictions unchanged.}
\mname outperforms baselines on alignment benchmarks such as Flask, HelpSteer, and BIGGen, with improvements of up to 7.5\% in Spearman correlation. Although reasoning-based evaluation methods provide more transparency by exposing intermediate reasoning steps, their reasoning is often shallow and fails to improve reliability, leading to inferior performance. In contrast, without explicit reasoning, \mname achieves comparable or superior results to these reasoning-based baselines. We further select instruction data with \mname method and achieve better performance on the AlpacaEval-2.0 benchmark than various baselines. Experiments on more downstream applications show the effectiveness of the \mname method.
\footnote{Our code is publicly available at \url{https://github.com/sustech-nlp/LAGER}.} 

\begin{figure*}[!tbp]
\centering
\includegraphics[width=1\textwidth]{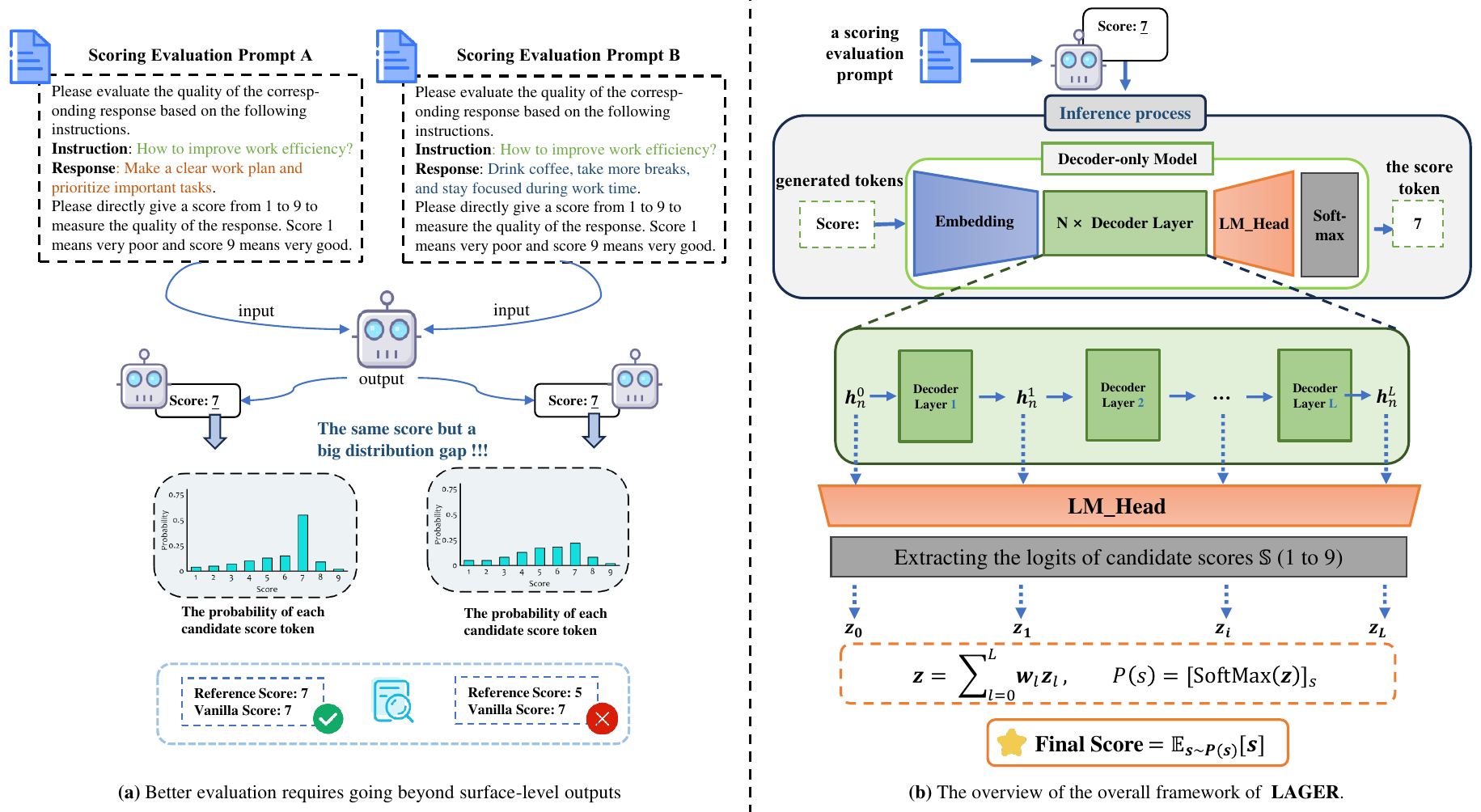}  
\caption{\textbf{(a)}: Vanilla scores may overlook meaningful distinctions by relying solely on the most probable score token. Better evaluation requires leveraging deeper signals beyond surface-level outputs. \textbf{(b)}: The framework of \mname. \mname fully considers candidate score probabilities and aggregates cross-layer logits, enabling LLMs to act as robust evaluators with probabilistic scoring for reliable assessment.}   
\label{fig:framework_overview}
\end{figure*}

\section{Preliminary}
A decoder-only LLM \( f \) consists of an embedding layer $f_{\text{embd}}$, \( L \) decoder layers $f_{\text{decoder}}$, and an output projection layer $f_{\text{unembd}}$. Given an input instruction \( x \), the model generates a response \( y \) through the following computation:
\begin{equation}
    y = f(x) = f_{\text{unembd}} \circ \underbrace{f_{\text{decoder}}^{(L)} \circ \cdots \circ f_{\text{decoder}}^{(1)}}_{\text{all decoder layers}} \circ f_{\text{embd}}(x),
\end{equation}
where $\circ$ denotes function composition, i.e., $(f \circ g)(x) = f(g(x))$. To predict the next token \( x_i \), the model typically relies on the hidden state produced by the final decoder layer, denoted as
\begin{equation}
    \mathbf{h}^{(L)}_i = f_{\text{decoder}}^{(L)} \circ \cdots \circ f_{\text{decoder}}^{(1)} \circ f_{\text{embd}}(x_{<i}),
\end{equation}
where \( x_{<i} = (x_1, \dots, x_{i-1}) \) represents the sequence of previous tokens. The output logits for token \( x_i \) are then computed as:
\begin{equation}
    \hat{\mathbf{z}}_i = f_{\text{unembd}}(\mathbf{h}^{(L)}_i).
\end{equation}
When incorporating a general LLM as the judge for a model response, the prompt template includes the evaluation task description \( x_d \), user instruction \( x_i \), the response to be evaluated \( x_r \), and the scoring criteria and output format requirement \( x_c \) \citep{g-eval,hdeval,hu2024llmbasedevaluatorsconfusingnlg}. We denote the full input as \( X = \{x_d, x_i, x_r, x_c\} \), and the set of candidate scores as \( \mathbb{S} \subset \mathcal{V} \), where \( \mathcal{V} \) is the model's vocabulary. Let \( \mathbf{h}^{(L)}_n \) denote the final-layer hidden state at the position where the score token is to be generated (e.g., after 'Score:'). The model produces output logits:
\[
\hat{\mathbf{z}}_n = f_{\text{unembd}}(\mathbf{h}^{(L)}_n),
\]
which are then passed through a softmax over the model's vocabulary \( \mathcal{V} \):
\begin{equation}
    P_{L}(t|X,y_{<n}) = \frac{\exp(\hat{\mathbf{z}}_n[t])}{\sum_{t^{'} \in \mathcal{V}} \exp(\hat{\mathbf{z}}_n[t'])}.
    \label{eq:softmax}
\end{equation}
The score token with the highest probability is the final score, referred to as the \textbf{vanilla score}:
\begin{equation}
    s_{L}^* = \arg\max_{s \in \mathbb{S}} P_{L}(s|X,y_{<n}).
    \label{eq:argmax_score}
\end{equation}
Figure~\ref{fig:framework_overview}(a) shows that vanilla score overlooks informative score distributions by relying solely on the top-probability score token, limiting its ability to distinguish response quality. More reliable and fine-grained evaluation requires the use of deeper internal representations beyond surface output.
\section{Methods} \label{sec:method}

\subsection{Motivation} \label{subsec:motivation}
\begin{wrapfigure}{r}{0.5\textwidth}
\vspace{-10pt}
\centering
    \includegraphics[width=1\linewidth]{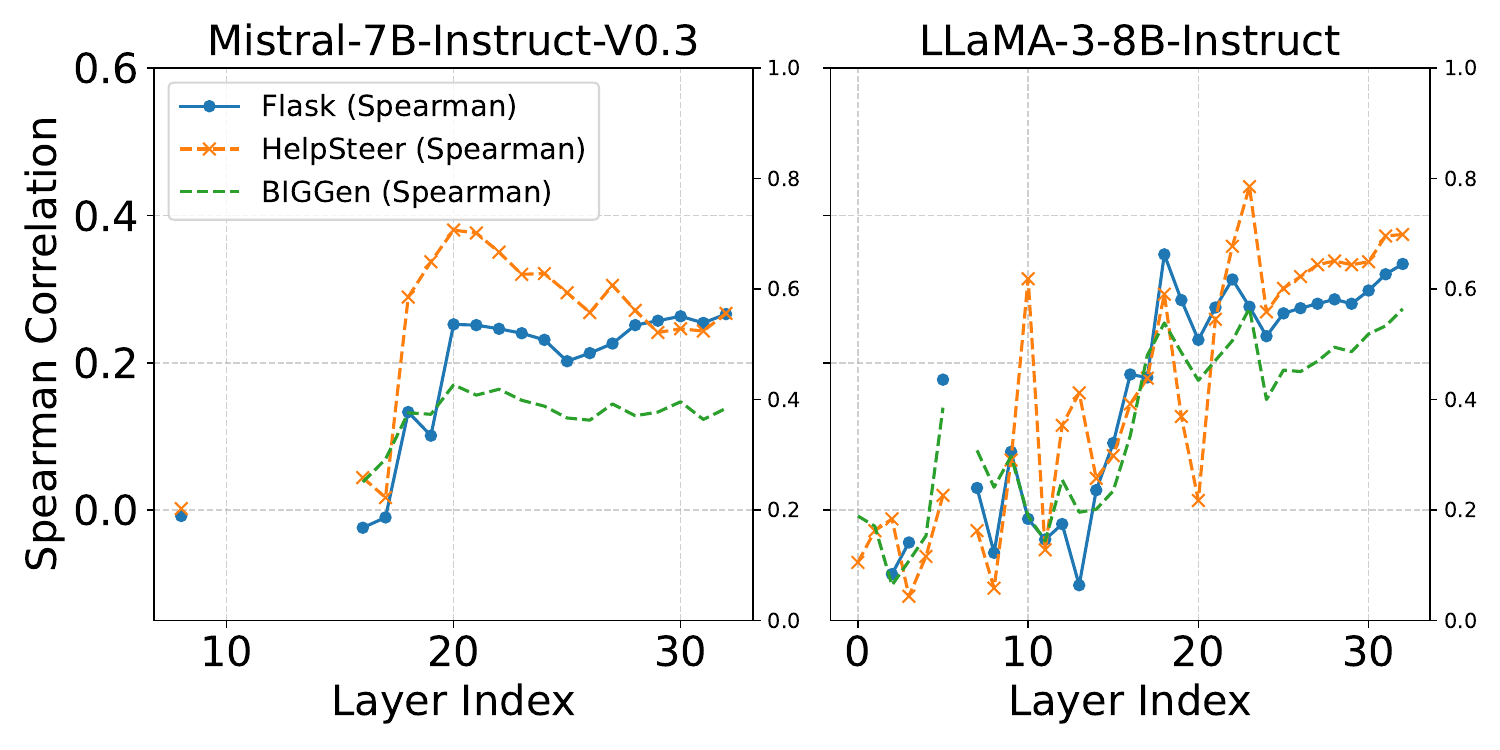}
    \caption{Agreement Between Human Ratings and Internal Layer Scores of Different Models.}
    \label{fig:layer_alys}
\vspace{-5pt}
\end{wrapfigure}
Hidden representations across different layers have been widely observed to exhibit distinct characteristics: the bottom layers focus more on local lexical information, while the middle and top layers focus more on semantic and global information\citep{wang-etal-2020-layer,yang2023gtrans,roh2024levi,transformerlayerspainters}.
Previous works~\citep{transformerlayerspainters,alabi2024hiddenspacetransformerlanguage} using techniques such as PCA and shared-space analysis have shown that middle-layer hidden states are highly consistent and interchangeable across language adaptations, suggesting a shared representation space. Thus, they can be directly transformed to logits with the shared LLM output unembedding layer $f_{\text{unembd}}$.

Previous works on LLM-as-a-Judge predominantly rely on the final-layer hidden state, while ignoring the intermediate representations across layers. Hence, we investigate the judge scores derived from hidden representations at different layers to better understand the layer-wise evaluation behavior. Specifically, the logits at the $l_{th} (l<L)$ layer can be computed as follows:
\begin{equation}
    \hat{\mathbf{z}}_l =f_{\text{unembd}}(f_{\text{decoder}}^{(l)} \circ \cdots \circ f_{\text{decoder}}^{(1)} \circ f_{\text{embd}}(x_{<n})) =f_{\text{unembd}}(\mathbf{h}^{(l)}_n),
\end{equation}
Following Equation~\ref{eq:softmax}, we can obtain the probability distribution $P_{l}(t|X,y_{<n})$over the vocabulary for each token, which allows us to compute the final judge score at the $l_{th}$ layer:
\begin{equation}
    s^*_l = \arg\max_{s \in \mathbb{S}} P_{l}(s|X,y_{<n}).
\end{equation}
Following the above setup, we conduct evaluations of different models' layer-wise scoring performance across multiple benchmarks and analyzed their alignment with human judgments. The results\footnote{For certain layers, if the calculated scores are identical across all samples, the Spearman correlation cannot be computed; such cases are shown as missing points in the figure.} in Figure~\ref{fig:layer_alys} show a consistent pattern across various benchmarks and models: \textbf{one or more intermediate layers yield better agreement with human evaluations than the final layer.}
% Inspired by our preliminary observations, we explore incorporating the hidden representations from selective layers to improve the judge score estimation.
% We propose the \mname framework to enhance the consistency with judgment from human experts while maintaining the generalization ability of the LLM evaluator, and achieve better alignment with human score distributions.

In this paper, we focus on point-wise LLM evaluators without reference response. Inspired by our preliminary observations, we explore incorporating the hidden representations from intermediate layers to improve the judge score estimation. We propose the \mname framework to enhance the consistency with judgment from human experts while maintaining the generalization ability of the LLM evaluator, and achieve better alignment with human score distributions.
\subsection{Score Estimation Based on LLM Hidden Representations} \label{sec:method_states}
For improved formality and clarity, we rewrite the LLM output unembedding layer $f_{\text{unembd}}$ as $\bm{W}_{\mathrm{unembd}}$, so $\hat{\bm{z}_{l}}$ can rewrite as:
\begin{equation}
    \hat{\bm{z}_{l}} = (f_{\text{decoder}}^{(l)} \circ \cdots \circ f_{\text{decoder}}^{(1)} \circ f_{\text{embd}}(x_{<n}))\bm{W}_{\mathrm{unembd}} =\mathbf{h}^{(l)}_n\bm{W}_{\mathrm{unembd}}.
\end{equation}
Based on the results in Figure~\ref{fig:layer_alys}, identifying consistently well-performing layers is challenging, as the layers exhibit varying levels of alignment with human judgments across different datasets. To address this, we first aggregate the logits corresponding to the set of candidate scores $\mathbb{S}$ from different layers using a set of weights $\bm{w} = \{w_0, w_1, \cdots, w_L\}$, where $w_0$ corresponds to the output of the embedding layer. The above procedure can be represented as:
\begin{equation}
    \hat{\bm{z}} = \sum_{i=0}^{L}w_i \hat{\bm{z}_{i}} = \sum_{i=0}^{L}w_i\mathbf{h}^{(i)}_n\bm{W}_{\mathrm{unembd}},
\end{equation}
following Equations~\ref{eq:softmax} and \ref{eq:argmax_score}, we can calculate a integer judge score. However, as illustrated in Figure~\ref{fig:framework_overview}(b), instead of relying on a single token, the full probability distribution over score candidates can be leveraged to capture richer evaluative information. This yields a probability distribution over the discrete score set, from which we compute a continuous, fine-grained score by taking the expected value. Therefore, we first extract the logits corresponding to all candidate score tokens from the full vocabulary logits and then aggregate them:
\begin{equation}
    \hat{\bm{z}}_{[\mathcal{M}]} = \sum_{i=0}^{L}w_i \hat{\bm{z}_{i}}_{[\mathcal{M}]} = \sum_{i=0}^{L}w_i[\mathbf{h}^{(i)}_n\bm{W}_{\mathrm{unembd}}]_{\mathcal{M}},
\end{equation}
where $\mathcal{M} = \{\mathrm{Tokenize}(s)|s\in\mathbb{S}\}$ denotes the set of tokenized sequences from $\mathbb{S}$. The final probability distribution of $\mathbb{S}$ is then obtained by applying a softmax over the aggregated logits:
\begin{equation}
    P(s) = \frac{\exp(\hat{\mathbf{z}}[s])}{\sum_{s^{'} \in \mathbb{S}} \exp(\hat{\mathbf{z}}[s'])},\quad s\in \mathbb{S}
\end{equation}
by taking the expectation over the score distribution, we can obtain the final fine-grained judge score:
\begin{equation}
    s^{*} = \mathbb{E}_{s \sim P(s)}[s] = \sum_{s \in \mathbb{S}}s\times P(s).
\label{eq:judge_logits}
\end{equation}
Normalizing before aggregation removes the relative scale information of logits and overemphasizes less important layers. We compare both settings in the ablation study (Section~\ref{sec:ablation_study}).
\subsection{Lightweight Training of Layer Weights}
We propose two types of layer weights $\bm{w}$. One is to apply average aggregation $w_l=1/(L+1)$ (denoted as \mname (w.o. tuning) in Table~\ref{table:psk_overall}), the other is to tune the lightweight $L+1$ parameters ($L$ is the number of transformer layers) with a small-scale validation set (denoted as \mname (w. tuning)). With a frozen backbone and minimal learnable parameters, we refer to the dataset as a validation set to distinguish it from finetuning-based LLM evaluators. \textit{To enhance the model's performance while aligning its predictions more closely with the distribution of human scores}, we adopt a combination of cross-entropy(CE) loss and mean absolute error (MAE) loss, balanced by a weighting hyperparameter $\alpha$.
\begin{equation}
    \begin{aligned}
    \mathcal{L}_{\text{Final}} &=\alpha \cdot \mathcal{L}_{\text{CE}} + (1-\alpha)\cdot \mathcal{L}_{\text{MAE}}
    \\&= \alpha\cdot \left( - \frac{1}{\mathcal{|B|}} \sum_{i=1}^{\mathcal{|B|}} \sum_{s=1}^{\mathbb{S}}\mathbb{I}(s=s_{\mathrm{truth}}^i) \log P_i(s) \right) + \left(1-\alpha\right)\cdot \left( \frac{1}{2\mathcal{|B|}} \sum_{i=1}^{\mathcal{|B|}}(s_i^*-s_{\mathrm{truth}}^i)^2 \right)
\end{aligned}
\end{equation}
where $\mathcal{B}$ is the batch of samples in a training step, and $s_{truth}^i$ is the human-annotated score for each sample. Please refer to Appendix~\ref{sec:app_training_layer_weights} for specific training settings and details. \textbf{It is important to note that the weights are tuned only once for each backbone and subsequently applied across all benchmarks and downstream tasks.} As the LLM remains frozen, no additional judge model is needed for scenarios like self-improvement. Our method is implemented without altering the model's next-token prediction process or logits, thereby offering plug-and-play capability. In contrast, finetuning-based methods require domain-specific judge models and suffer from limited generalization due to scarce labeled data.

\subsection{Discussion: Applicability under Restricted Access Settings}
While \mname relies on accessing the intermediate layers of the LLM, it can naturally adapt to more restricted scenarios. For instance, when only final-layer logits are available (e.g., API-based models), we can extract the logits corresponding to score tokens, apply softmax, and compute the expected score (this method is proposed in G-Eval~\citep{g-eval}).  We refer to this setup as \textbf{expectation score} (\textbf{E-Score}). The evaluation uses the \textbf{vanilla score}, based on the most likely token, when only the final token prediction is observable. Both settings can be viewed as simplified special cases within our framework. We treat both scores as comparable baselines in our experiments. \textit{Nonetheless, powerful open-source models still remain the mainstream for large-scale evaluation due to their lower cost and performance comparable to closed-source models}~\citep{rewardbench,gu2024surveyllmasajudge,malik2025rewardbench2advancingreward}, {making \mname broadly applicable and easily extensible to future models}.

\section{Experiments}

\subsection{Experiments Setup} \label{sec:exp_setup}
\textbf{Benchmarks.} Due to potential preference leakage~\citep{preferenceleakage} between models, we evaluate our method on human-annotated datasets for reliable results. We chose three diverse point-wise benchmarks to ensure a comprehensive evaluation: \textbf{Flask}~\citep{flask},  \textbf{HelpSteer}~\citep{wang2023helpsteermultiattributehelpfulnessdataset}, and \textbf{BiGGen Bench}\footnote{\href{https://huggingface.co/datasets/prometheus-eval/BiGGen-Bench-Results}{https://huggingface.co/datasets/prometheus-eval/BiGGen-Bench-Results}} ~\citep{prometheus2}. In these benchmarks, human-annotated scores range from 1 to 5. Please refer to Appendix~\ref{sec:app_bench} for the details. 

\textbf{Models.} We experiment with multiple backbone models of varying sizes and from different families, including Mistral-7B-Instruct-v0.3~\citep{mistral7b}, InternLM3-8B-Instruct~\citep{cai2024internlm2}, LLaMA3.1-8B-Instruct~\citep{llama3herdmodels}, Qwen-2.5-14B-Instruct~\citep{qwen2025qwen25technicalreport}, Mistral-Small-24B-Instruct and LLAMA-3.3-70B-Instruct. 

\textbf{Baselines.} We compare \mname with three baselines: \textbf{GPTScore}, \textbf{vanilla score (VScore)} and \textbf{expectation score (E-Score)}. At the same time, we compare the commonly used API-based model \textbf{GPT-4o-mini}. We also experiment with methods that explicitly train the LLM backbones, namely \textbf{TIGERScore-7B} and \textbf{Prometheus2-7B}. While \mname does not modify the backbone models, these methods are included solely for reference and are not directly compared. Please refer to Appendix~\ref{app:baselines} for specific details about the baselines.

\textbf{Evaluation Metrics.} Following previous works~\citep{gu2024surveyllmasajudge,benchmarkingfoundationmodelslanguagemodelasanexaminer,aligninghumanjudgmentrole}, we use the commonly adopted \textbf{Pearson} and \textbf{Spearman} correlations to measure the consistency between LLM-predicted scores and human annotations. See Appendix~\ref{sec:app_metrics} for further details. 
\begin{table}[!t]
\centering
\caption{Spearman Correlation Results: \mname vs. Baselines on Flask, HelpSteer, and BiGGen Bench. \textit{These
results are statistically significant, with p-values less than 1e-5}. See Table~\ref{table:pearson results} for the full Pearson correlation results.} 
\resizebox{\textwidth}{!}{
\begin{tabular}{lcccccccc}
\toprule
\multirow{2}{*}{\textbf{Model}} &  & \multicolumn{2}{c}{Flask} & \multicolumn{2}{c}{HelpSteer} & \multicolumn{2}{c}{BIGGen Bench} & \multirow{2}{*}{Average} \\ \cline{3-8}
 &  & Direct & Reasoning & Direct & Reasoning & Direct & Reasoning & \\ \hline
\multicolumn{9}{c}{\cellcolor[HTML]{DAE8FC}\textit{\textbf{Fine-tuned Models}}} \\ \hline
\textbf{TIGERscore-7B} &  & - & 0.175 & - & 0.118 & - & 0.171 & 0.155 \\
\textbf{Prometheus2-7B} &  & - & 0.413 & - & 0.514 & - & 0.367 & 0.431 \\ \hline
\multicolumn{9}{c}{\cellcolor[HTML]{DAE8FC}\textit{\textbf{Close-source Model via API}}} \\ \hline
\textbf{GPT-4o-mini} &  &  &  &  &  &  &  &  \\ 
Vscore &  & 0.526 & 0.535 & 0.482 & 0.535 & 0.534 & 0.509 & 0.520 \\
E-Score &  & 0.579 & 0.561 & 0.500 & 0.541 & 0.573 & 0.530 & 0.547 \\ \hline
\multicolumn{9}{c}{\cellcolor[HTML]{DAE8FC}\textit{\textbf{Open-source Models}}} \\ \hline
\textbf{Mistral-7B-Instruct-v0.3} &  &  &  &  &  &  &  &  \\ 
GPTScore &  & 0.258 & - & 0.209 & - & 0.183 & - & 0.217 \\
Vscore &  & 0.266 & 0.269 & 0.267 & 0.364 & 0.138 & 0.280 & 0.264 \\
E-Score &  & 0.239 & 0.279 & 0.296 & \textbf{0.380} & 0.185 & 0.283 & 0.277 \\
\mname(w.o tuning) &  & 0.338 & 0.295 & 0.401 & 0.377 & 0.353 & 0.329 & 0.349 \\
\mname(w. tuning) & \textbf{} & \textbf{0.347} & \textbf{0.298} & \textbf{0.403} & 0.376 & \textbf{0.357} & \textbf{0.333} & \textbf{0.352} \\ \hline
\textbf{LLaMA3.1-8B-Instruct} &  &  &  &  &  &  &  &  \\ 
GPTScore &  & 0.061 & - & -0.022 & - & -0.162 & - & -0.041 \\
Vscore &  & 0.334 & 0.429 & 0.374 & 0.518 & 0.273 & 0.390 & 0.386 \\
E-Score &  & 0.386 & 0.446 & 0.464 & \textbf{0.525} & 0.352 & 0.403 & 0.429 \\
\mname(w.o tuning) &  & 0.472 & 0.456 & \textbf{0.520} & 0.524 & 0.475 & 0.443 & 0.482 \\
\mname(w. tuning) & \textbf{} & \textbf{0.477} & \textbf{0.460} & 0.515& 0.524& \textbf{0.482}& \textbf{0.444}& \textbf{0.484}\\ \hline
\textbf{InternLM3-8B-Instruct} &  &  &  &  &  &  &  &  \\ 
GPTScore &  & -0.087 & - & -0.062 & - & -0.257 & - & -0.135 \\
Vscore &  & 0.423 & 0.449 & 0.388 & 0.425 & 0.374 & 0.441 & 0.417 \\
E-Score &  & 0.515 & 0.472 & 0.453 & 0.430 & 0.470 & 0.470 & 0.468 \\
\mname(w.o tuning) &  & 0.449 & 0.468 & 0.426 & 0.429 & 0.374 & 0.469 & 0.436 \\
\mname(w. tuning) & \textbf{} & \textbf{0.545} & \textbf{0.489} & \textbf{0.515} & \textbf{0.474} & \textbf{0.507} & \textbf{0.490} & \textbf{0.501} \\ \hline
\textbf{Qwen-2.5-14B-Instruct} &  &  &  &  &  &  &  &  \\ 
GPTScore &  & 0.001 & - & -0.014 & - & -0.142 & - & -0.052 \\
Vscore &  & 0.547 & 0.537 & 0.420 & 0.423 & 0.458 & 0.461 & 0.474 \\
E-Score &  & 0.579 & 0.555 & \textbf{0.447} & 0.452 & 0.502 & 0.457 & 0.499 \\
\mname(w.o tuning) &  & 0.572 & 0.567 & 0.433 & \textbf{0.473} & 0.503 & 0.507 & 0.509 \\
\mname(w. tuning) & \textbf{} & \textbf{0.612}& \textbf{0.572} & 0.443& 0.472& \textbf{0.567}& \textbf{0.524}& \textbf{0.531} \\ \hline
\textbf{Mistral-Small-24B-Instruct} &  &  &  &  &  &  &  &  \\ 
GPTScore &  & 0.016 & - & -0.008 & - & -0.147 & - & -0.046 \\
Vscore &  & 0.528 & 0.505 & 0.420 & 0.459 & 0.542 & 0.533 & 0.498 \\
E-Score &  & 0.577 & 0.532 & 0.442 & 0.486 & 0.585 & 0.555 & 0.530 \\
\mname(w.o tuning) & \textbf{} & 0.591 & \textbf{0.542} & 0.452 & 0.485 & 0.589 & 0.562 & 0.537 \\
\mname(w. tuning) & \textbf{} & \textbf{0.596}& \textbf{0.542} & \textbf{0.449} & \textbf{0.487} & \textbf{0.598} & \textbf{0.566} & \textbf{0.540} \\ \hline
\textbf{LLAMA-3.3-70B-Instruct} &  &  &  &  &  &  &  &  \\ 
GPTScore &  & 0.042 & - & 0.014 & - & -0.193 & - & -0.046 \\
Vscore &  & 0.518 & 0.567 & 0.435 & 0.494 & 0.559 & 0.539 & 0.519 \\
E-Score &  & 0.464 & 0.540 & 0.444 & 0.488 & 0.530 & 0.506 & 0.495 \\
\mname(w.o tuning) &  & 0.610 & 0.598 & \textbf{0.506} & \textbf{0.520} & 0.597 & \textbf{0.585} & 0.569 \\
\mname(w. tuning) & \textbf{} & \textbf{0.611} & \textbf{0.598}& 0.504 & 0.519 & \textbf{0.602} & 0.584 & \textbf{0.570} \\ \bottomrule
\end{tabular}}
    \label{table:psk_overall}
\vspace{-20pt}
\end{table}

\textbf{Evaluation Details.} We adopt greedy decoding for its stability and consistency in evaluation~\citep{LLM-eval}. %Greedy decoding selects the most probable token at each step, minimizing randomness in the inference process and ensuring consistency and reproducibility of the results. 
To comprehensively assess the effectiveness of our method,  we conduct evaluations under two different conditions:
\textbf{1) Direct:} Directly providing a score during evaluation without reasoning.
\textbf{2) Reasoning:} Previous work~\citep{g-eval,wang2024dhpbenchmarkllmsgood,hu2024llmbasedevaluatorsconfusingnlg} shows that reasoning improves judgment quality, with reasoning-first setups performing best\citep{feedback-first,murugadoss2024evaluatingevaluatormeasuringllms,gu2024surveyllmasajudge}. Thus, we adopt a reasoning-first approach in this study.
The prompt template used in our work follows the standard configuration widely adopted in prior studies~\citep{prometheus2,g-eval,instructscore}, please refer to Appendix~\ref{sec:app_prompt_template} for the prompt templates. Unless otherwise specified, we highlight the best results in each table using \textbf{bold}, and denote the second-best with \underline{underlining}.

\subsection{Main Results} \label{sec:main_results}
Tables~\ref{table:psk_overall} report Spearman results of \mname compared to baselines across three benchmarks. Section~\ref{sec:qwen2.5-family} further analyzes the performance variation of \mname and baseline methods across different model scales, under the setting where models from the Qwen2.5 family are used as frozen backbones, and compares their results on the same set of benchmarks. Specifically, \mname (w. tuning) achieves near-best performance on these benchmarks and demonstrates strong stability, so it serves as the main approach in our work.

\paragraph{The Effectiveness and Scalability of LAGER.} \mname with tuning consistently outperforms all non-training baselines across diverse LLM backbones on all three benchmarks. This highlights the importance of leveraging internal logits and output distributions for accurate scoring. Although Mistral-Small-24B-Instruct and Qwen2.5-14B-Instruct underperform GPT-4o-mini on Vscore, they achieve comparable results when using \mname. GPTScore performs poorly due to its reliance on generation probabilities, which overlook the semantic and logical quality. Due to its focus on error detection, the fine-tuned TIGERscore-7B shows relatively limited performance. Compared to the fine-tuned Prometheus2-7B, \mname also enables InternLM3-8B-Instruct and LLaMA3.1-8B-Instruct, two models of comparable size, to surpass it even though they initially have weaker performance on Vscore. Even without tuning, \mname outperforms E-Score in 5 out of 6 backbone models, with an average improvement of 4.5\%. In the few cases where it underperforms, the performance gap is within 3\%. Moreover, it consistently outperforms V-Score. 
\paragraph{The Robustness of LAGER.} The average E-Score performance of models like Qwen-2.5-14B-Instruct, LLAMA-3.3-70B-Instruct, and Mistral-7B-Instruct-v0.3 is sometimes lower than VScore. Specifically, the average E-score performance of LLAMA-3.3-70B-instruct across all benchmarks is 2.4 points lower than Vscore. In contrast, \mname with tuning achieves almost the best performance across all settings and benchmarks, consistently outperforms Vscore. \mname can closely match human score distribution and incorporate more fine-grained information. More details are discussed in Section~\ref{sec:analysis}. Therefore, we primarily recommend the \mname with tuning approach and conduct subsequent analyses based on it. The performance gains of LAGER stem not from large-scale fine-tuning, but from a more efficient and structured utilization of the model's internal representations.
\paragraph{Reasoning Is Not Always Better.} The richness of information aggregation increases gradually from Vscore, E-Score to \mname(with tuning). Figure~\ref{fig:direct_vs_reasoning} shows that reasoning leads to an improvement in Vscore performance, a divergence in E-Score performance (with half of the models improving and the other half declining), and consistently weaker results for \mname (with tuning) under the reasoning setting compared to direct. We speculate that this phenomenon may be due to reasoning potentially causing the model to become overconfident \citep{yang2024trustllmsmitigateoverconfidence,wagner2024blackboxuncertaintyquantificationmethod}, which in turn leads to inaccurate evaluation scores. This is also consistent with the findings of previous work~\citep{ge-etal-2025-backdoors,guo2024activedormantattentionheadsmechanistically,li2024measuring}: As the reasoning steps progress, the model's focus on the original input and the text to be evaluated gradually diminishes, shifting more toward its self-generated reasoning trajectory.
\begin{table}[ht]
\centering
\caption{Ablation Study Results on FLASK and HelpSteer Datasets Evaluated by Spearman Corr.}
\resizebox{0.8\linewidth}{!}{
\begin{tabular}{cccccc|cccc}
\toprule
\multicolumn{6}{c|}{\textbf{Strategies For \mname}} & \multicolumn{2}{c}{\textbf{InternalLM}} & \multicolumn{2}{c}{\textbf{Mistral}} \\ \hline
ID & Exp.& Max.&Logits agg. & Prob. agg. & Tuning & Flask & HelpSteer & Flask & HelpSteer \\ 
\toprule
\textcircled{1}& $\checkmark$ & $\times$ & $\checkmark$ & $\times$ & $\checkmark$ & \textbf{0.545} & \textbf{0.515} & \textbf{0.347} & \textbf{0.403} \\
\textcircled{2}& $\checkmark$ & $\times$ & $\times$ & $\checkmark$ & $\checkmark$ & 0.452 & 0.448 & 0.299 & 0.344 \\
\textcircled{3}& $\times$ & $\checkmark$ &$\checkmark$ & $\times$ &   $\checkmark$ & 0.373 & 0.374 & 0.268 &  0.264\\
\textcircled{4} &$\times$ & $\checkmark$ &$\times$ &$\checkmark$ &   $\checkmark$ &0.367 &0.379&0.268 &0.267 \\
\textcircled{5} & $\checkmark$ & $\times$ & $\checkmark$ & $\times$ & $\times$ & 0.449 & 0.426 & \underline{0.338} &\underline{0.401} \\
\textcircled{6} & $\checkmark$ & $\times$ & $\times$ & $\checkmark$ & $\times$ &  0.442 & 0.445 & 0.306 &  0.358\\
\textcircled{7} & $\times$ & $\checkmark$ &$\checkmark$ & $\times$ &  $\times$ & 0.379 & 0.367 & 0.265 &  0.264\\
\textcircled{8} & $\times$ & $\checkmark$ &$\times$ &$\checkmark$ &   $\times$ & 0.364 & 0.363 & 0.268 &  0.267\\
\textcircled{9} & $\checkmark$ & $\times$ &$\times$ &$\times$ & $\times$ & \underline{0.515} & \underline{0.453} & 0.239 & 0.296 \\
\textcircled{\small{10}} & $\times$ & $\checkmark$ & $\times$ & $\times$ & $\times$ & 0.423 & 0.388 & 0.266 & 0.267 \\ \bottomrule
\end{tabular}}
\label{table:ablation}
\end{table}
\subsection{Ablation Study} \label{sec:ablation_study}
To better understand each design choice of \mname, we conduct a comprehensive ablation study on the direct setup using Mistral-7B-Instruct-v0.3 and InternLM3-8B-Instruct. Table~\ref{table:ablation} shows the results of the Spearman correlation. The expectation (Exp.) score and the maximum (Max.) score are alternative designs. Given the score distribution of a certain layer, we use Eqn.\ref{eq:judge_logits} to calculate the expected score or use the score with the highest probability. Logits agg. and prob. agg. are two alternative designs, aggregating information from all hidden layers at the logits level or output distribution level. Removing logits agg. and prob. agg. indicates that we do not use internal states. It is observed that the full configuration (Exp. score + Logits agg. + Tuning) achieves the best performance across all settings, with a maximum Spearman correlation of 0.545; Fine-tuning brings up to +0.10 improvement; expectation scoring outperforms maximum scoring (up to +0.17); and logits aggregation surpasses probability aggregation (up to +0.07). Multi-layer integration is generally effective, especially for Mistral; however, in untuned settings, InternalLM occasionally performs better without aggregation, indicating model-specific differences.
\section{Analyses} \label{sec:analysis}
\subsection{Understanding Internal States}
\begin{figure}[h]
    \centering
    \includegraphics[width=1\linewidth]{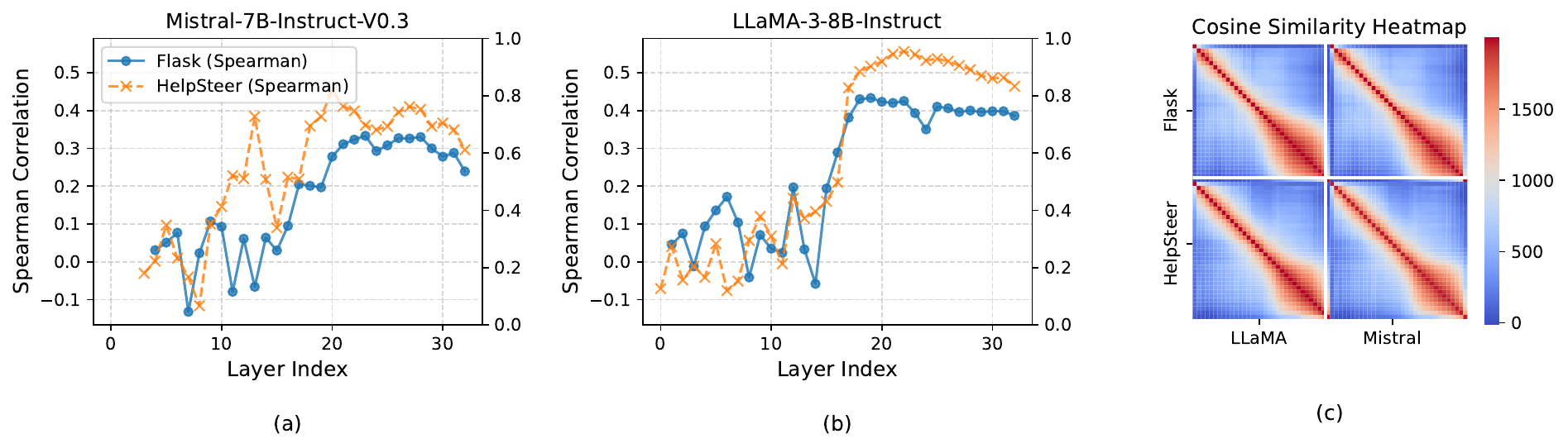}
    \caption{Analysis of Llama3.1-8B-Instruct and Mistral-7B-Instruct-V0.3 using HelpSteer and Flask. \textbf{(a)} and \textbf{(b)} illustrate the performance of E-scores computed from different layers. \textbf{(c)} depicts the average cosine similarity heatmap of hidden states across layers.}
    % along with the layer-specific weights learned during training.
    \label{fig:layer-wise}
\vspace{-10pt}
\end{figure}
Having observed that internal states contribute significantly to the success of \mname, we now aim to understand how internal states in each layer contribute to the judgment score. 
Specifically, we apply \mname to each layer of the backbone LLM, namely using the expected score from the output distribution of each layer, and report the consistency with human scores.
Figure~\ref{fig:layer-wise} illustrates the results of the Mistral-7B-Instruct-v0.3 and LLaMA-3.1-8B-Instruct backbones. We observe the following:
(1)\textbf{The bottom layers exhibit low or even negative correlation with human scores}, indicating their limited utility in evaluation tasks. 
(2)The middle-upper layers (approximately layers 20 to 30) show the highest correlation with human scores, suggesting \textbf{they carry the most informative representations for judgment}. However, there is a drop in correlation at the topmost layer, implying that certain judgment-relevant signals may be diluted or lost in the final transformation stages. Thus, relying solely on the top layer is insufficient for precise evaluation.

To better understand the differences between internal and top layers, we visualize the cosine similarity between hidden states across layers at the position of the score token using the Flask and HelpSteer datasets. Based on Figure~\ref{fig:layer-wise}(c), the similarity between layers generally decreases as the distance between layers increases. For the bottom layers, the representation changes rapidly with increasing layer number. In contrast, the representation changes relatively slowly for the middle-upper layers, and the representations across these layers remain similar over a broader range. The topmost layer exhibits low similarity to its neighboring layers, displaying a distinct pattern. 

Based on the above observations, we believe that: \textbf{The middle-upper layers exhibit relatively consistent representations and stronger alignment with human scores}, indicating that these layers may deserve more attention. Meanwhile, \textit{the topmost layer shows comparatively weaker judgment ability, likely because its focus on next-token prediction leads to the loss of fine-grained evaluative information}. Instead of searching for optimal layers, which vary across LLM backbones and are hard to identify, \mname directly learns a weighted combination of all layers. This approach is both lightweight and effective, requiring only a small validation set and transferring well across benchmarks (Section~\ref{sec:main_results}, Appendix~\ref{sec:EU}, Appendix~\ref{sec:selfaware}) and tasks (Section~\ref{sec:data_filering}, Appendix~\ref{sec:app_lager}).

\subsection{LAGER Yields More Human-Aligned and Fine-Grained Score Distributions}
\begin{wrapfigure}{r}{0.4\textwidth}
\vspace{-20pt}
\centering
    \includegraphics[width=0.9\linewidth]{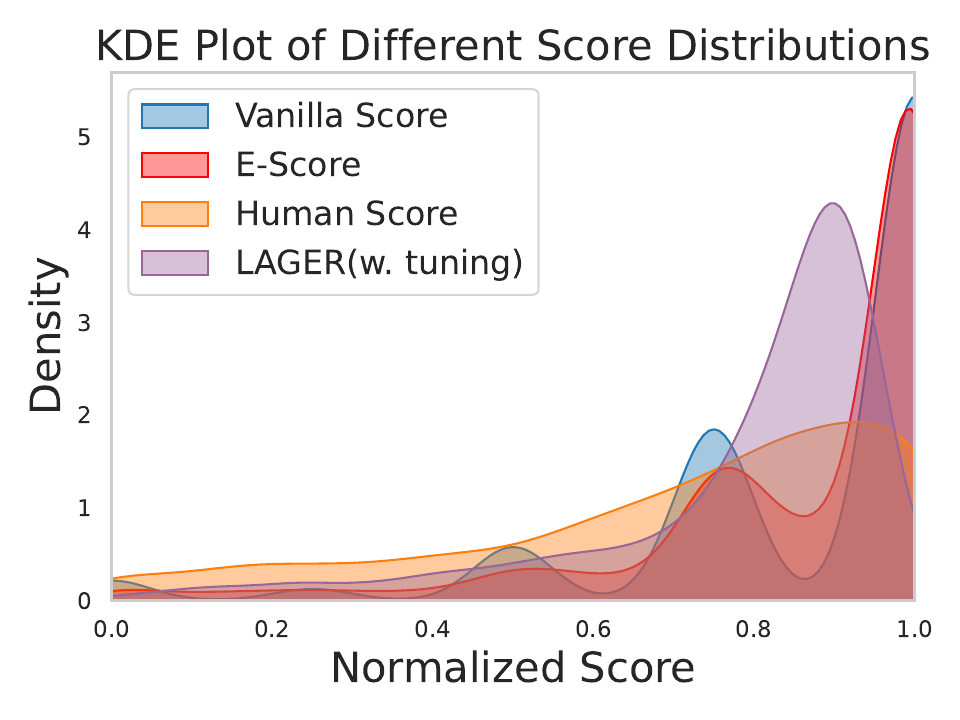}
    \caption{The KDE Plot of Different Score Distributions, evaluated on Flask with LLaMA-3.1-8B-Instruct backbone.}
    \label{fig:estimated-kernel-density}
\vspace{-5pt}
\end{wrapfigure}
To better understand how \mname improves over both the VScore and the E-Score, we evaluate LLaMA-3.1-8B-Instruct on the Flask dataset to obtain the Vanilla Score, E-Score, and \mname, and visualize the min-max normalized scores and human annotations using Gaussian kernel density estimation (KDE). Details of the KDE visualization procedure can be found in Appendix~\ref{sec:app_kde}. Figure~\ref{fig:estimated-kernel-density} visualizes the score distributions. Compared to E-Score and VScore, \mname yields a distribution that more closely matches human judgments, with noticeably higher overlap. \mname mitigates the high-score bias of VScore and E-Score by \textbf{shifting the score distribution to the left and producing more evenly distributed scores.}

\begin{wraptable}{r}{0.4\textwidth}
\vspace{-15pt}
\caption{Distribution distance of VScore and \mname to human score, evaluated on Flask with LLaMA-3.1-8B-Instruct backbone.}
\centering
\resizebox{0.4\textwidth}{!}{
\begin{tabular}{@{}lcc@{}}
\toprule
LLaMA-3.1-8B-Instruct & $\mathrm{D_{KL}}(\downarrow)$ & $\mathrm{MSE}(\downarrow)$ \\ \midrule
VScore & 0.312 & 0.112 \\
E-Score & 0.102 & 0.092 \\
\mname(w. tuning) & \textbf{0.087} & \textbf{0.060} \\ \bottomrule
\end{tabular}}
\label{tab:kl_and_mse}
\vspace{-20pt}
\end{wraptable}
We also quantitatively compare \mname, E-Score, and VScore by evaluating the Kullback-Leibler (KL) divergence and the mean squared error (MSE). The KL divergence of a candidate score $S$ is computed as $\mathrm{D}_{\mathrm{KL}}(\tilde{p}_{hs}||\tilde{p}_{s})$ where $\tilde{p}_{hs}$ and $\tilde{p}_{s}$ represent the human score distribution and the candidate score distribution, respectively. The results in Table~\ref{tab:kl_and_mse}, demonstrate that \mname more accurately approximates the distribution of human-annotated scores.

\subsection{From Tiny to Huge: LAGER Brings Improvements Across Model Scales}
To further investigate whether \mname can enhance human alignment across a broader range of model scales, we conducted experiments using three benchmarks on instruct models of varying sizes from the Qwen2.5 family, including 0.5B, 1.5B, 3B, 7B, 14B, 32B and 72B. 

\begin{figure}
    \centering
    \includegraphics[width=1\linewidth]{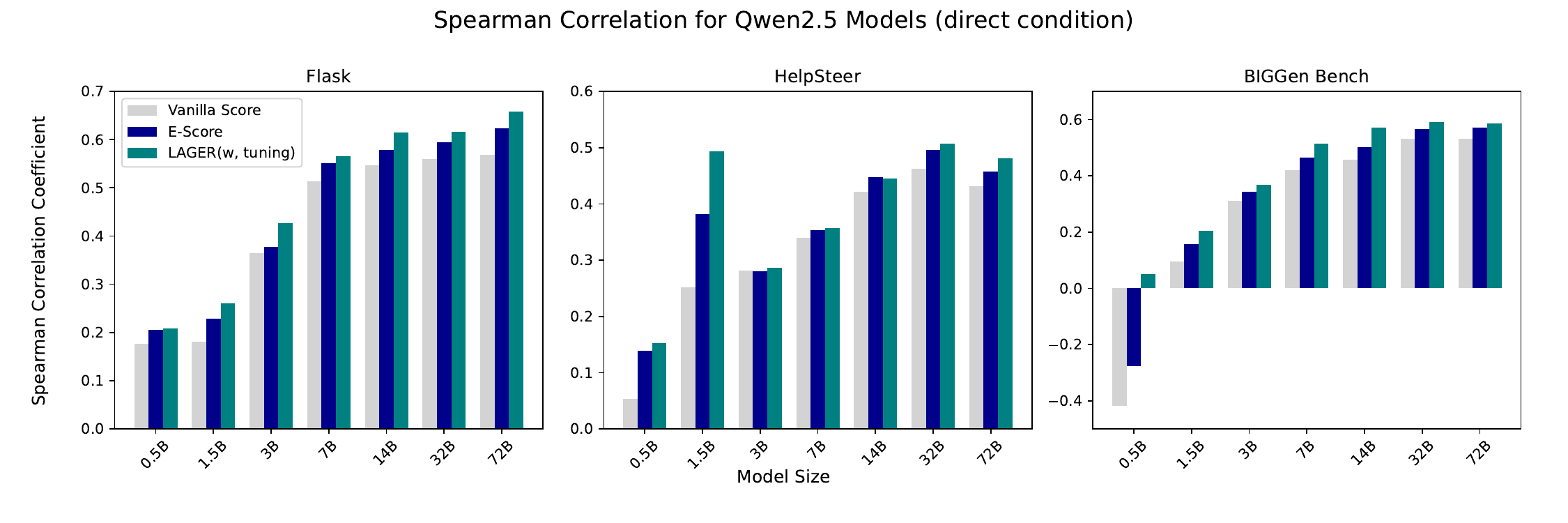}
    \caption{Comparison of Spearman Results for Qwen2.5 Models Across FLASK, HelpSteer and BIGGen Bench Using Vanilla Score, E-Score, and \mname(w. tuning) (Direct Condition).}
    \label{fig:qwens-direct}
\end{figure}

The Figure~\ref{fig:qwens-direct} and Figure~\ref{fig:qwens-reasoning} show the alignment effectiveness of Vanilla Score, E-Score, and \mname (w. tuning) across Qwen2.5 models from 0.5B to 72B. Both direct and reasoning conditions show a mostly upward trend in Spearman correlation coefficients as model size increases across Flask, Helpsteer, and BIGGen datasets, with \mname leading (e.g., 0.658 in Flask for direct, 0.598 for reasoning). When the model size exceeds 14B, \mname achieves greater performance gains over the backbone model than the VScore computed using GPT-4o-mini. Notably, on the BIGGen dataset, \mname shows a significant improvement on the 0.5B model scale, elevating the Spearman correlation from a negative value (-0.4 for the Vanilla Score) to a positive 0.1, marking a substantial gain compared to the baseline. These results demonstrate its effectiveness across different model scales and datasets, and the benefits brought by \mname are scale-transferable and synergistically amplified as model capacity grows.
% The results in the figure~\ref{fig:qwens-direct} and figure~\ref{fig:}demonstrate the alignment effectiveness of \mname across Qwen2.5 models of varying scales. From the smallest 0.5B model to the massive 72B model, all three scoring methods—vanilla Score, E-Score, and \mname (w. tuning), exhibit a consistent and steady improvement as the model size increases. This strong upward trend not only confirms the generalizability and effectiveness of \mname in aligning models with human preferences but also highlights its potential for deployment in resource-constrained settings. Therefore, the benefits brought by \mname are scale-transferable and synergistically amplified as model capacity grows.
\label{sec:qwen2.5-family}

\section{Applications of LAGER}  \label{sec:applications}

\subsection{Instruction Data Selection} \label{sec:data_filering}

Instruction data selection~\citep{instructionmininginstructiondata,longest-responses} involves choosing a subset of high-quality data from a given instruction dataset to enhance both the efficiency and performance of the tuning process. In this experiment, we use \mname as a scoring metric to select a 10\% subset from the alpaca-cleaned-52k dataset\footnote{\href{https://huggingface.co/datasets/yahma/alpaca-cleaned}{https://huggingface.co/datasets/yahma/alpaca-cleaned}} \citep{alpaca}. 
We compare our subset, Highest \mname, with four baselines on finetuning the LLaMA3-8B-base model: Longest instructions, Longest responses~\citep{longest-responses}, Highest VScore, and SuperFiltering \citep{superfiltering}. SuperFiltering is an explicitly designed instruction data filtering method. For methods that require a backbone LLM, we utilize LLaMA3.1-8B-Instruct for a fair comparison. More details are in Appendix \ref{sec:app_sft}.

\begin{table}
\caption{The Performance of LLaMA3-8B-Base Fine-tuning with Instruction Data Subsets on AlpacaEval 2.0: Length-Controlled (LC) win rate against GPT-4-1106-Preview (805 Test Examples), evaluated with GPT-4o-mini.}
\centering
\fontsize{16}{20}\selectfont
\resizebox{0.6\linewidth}{!}{
\begin{tabular}{ccccc}
\toprule
\multirow{2}{*}{\textbf{Filtering criteria}} & \multirow{2}{*}{\textbf{Ratio}}& & \multicolumn{2}{c}{\textbf{AlpacaEval-2.0}} \\ \cline{4-5} 
 & & & LC win rate & average length \\ \midrule
No filtering & 100\% && 7.73 & 1070 \\
Longest instructions & 10\% && 6.92 & 946 \\
Highest VScore & 10\% && 9.42 & 1206 \\
Highest E-Score & 10\% && 11.50 & 1282 \\ 
SuperFiltering & 10\% && 10.69 & 1367 \\
Longest responses & 10\% && \underline{11.82} & 1506 \\
Highest \mname(w.o. tuning)& 10\% && 11.77 & 1325 \\
Highest \mname(w. tuning)& 10\% && \textbf{12.65} & 1248 \\ 
\bottomrule
\end{tabular}}
\label{table:sft}
\vspace{-15pt}
\end{table}

As shown in Table~\ref{table:sft}, our method with tuning achieves the best performance, outperforming the second-best method, the Longest responses, and the explicitly designed method, SuperFiltering, by 0.83 and 1.96 LC win rate, respectively. Even without tuning, our method outperforms most baselines, improving the VScore by a 2.35 LC win rate. And the highest \mname(w tuning) has relatively shorter average lengths than most baselines. This indicates that our tuning model does not achieve better scores by generating longer responses, but by learning to follow instructions more effectively.

% . It outperforms the second-best method, Longest Responses, Highest \mname(w.o. tuning) and the explicitly designed method, SuperFiltering, by 3.58, 3.63 and 4.71 LC win rate, respectively. Additionally, the generated responses of 

% When comparing the generated lengths of different methods, we observe  

% Additionally, the generated responses have relatively shorter average lengths, indicating that under our scoring standards, the internal score alone is sufficient to filter out high-quality and appropriately concise responses. The results show that \textbf{when Direct Score fails to distinguish sample quality, \mname more effectively leverages the score distribution, assigning lower scores to low-quality data}. This further highlights the extensibility of the internal score.
% Summarizing the characteristics of our method from data selection.

% \subsection{Score-based Classification Tasks}
% To demonstrate the broader applicability of \mname, we evaluate its effectiveness in scenarios beyond merely assessing the quality of responses.

% Can \mname not only accurately evaluate the quality of responses but also showcase their unique advantages in other scenarios, thereby demonstrating their broader applicability and value?
% We chose two scenarios different from quality evaluation to verify the advantages of \mname compared to Drect Score:

In addition to the application mentioned above, \mname can be applied to other scoring-based tasks, such as \textbf{LLM emotional understanding} and \textbf{LLM Recognizing its Knowledge Boundaries}. \textbf{\mname shows strong transferability, even when there is a mismatch between the task's scoring range and that of the data used for training.}
Please refer to Appendix \ref{sec:EU} and Appendix \ref{sec:selfaware} for detailed discussions.

\section{Conclusion}

In this work, we propose \mname, an effective LLM-as-a-judge framework to estimate the judge score from the
logits information from different LLM layers. \mname offers training-free and tunable options, allowing flexible integration with various LLMs to deliver fine-grained judgment services. Unlike previous methods that explicitly update LLM backbones, \mname provides a lightweight tuning solution that only trains weights on a small validation dataset, and the learned weights can be transferred across tasks. Experiments on three comprehensive alignment benchmarks demonstrate the effectiveness of \mname. 
Specifically, \mname outperforms all baselines that do not explicitly train LLM backbones. Additionally, \mname demonstrates strong performance in improving alignment with human annotations across models of varying scales and families. Even without explicit reasoning, \mname outperforms various reasoning-based baselines, including those specifically trained for reasoning tasks, while requiring fewer generation tokens and offering higher efficiency. \mname is quite general and can be applied to various use cases, including but not limited to instruction data selection, recognizing its knowledge boundaries and emotional understanding.

\section*{Acknowledgements}

This project was supported by National Natural Science Foundation of China (No. 62306132), Guangdong Basic and Applied Basic Research Foundation (No. 2025A1515011564), Natural Science Foundation of Shanghai (No. 25ZR1402136). We thank the anonymous reviewers for their insightful feedback on this work.

% \clearpage
% \newpage

% In the unusual situation where you want a paper to appear in the
% references without citing it in the main text, use \nocite
% \nocite{langley00}
% \newpage

\small
{
\bibliography{custom}
\bibliographystyle{neurips_2024}
}

\newpage
% \include{check_list}
%%%%%%%%%%%%%%%%%%%%%%%%%%%%%%%%%%%%%%%%%%%%%%%%%%%%%%%%%%%%%%%%%%%%%%%%%%%%%%%
%%%%%%%%%%%%%%%%%%%%%%%%%%%%%%%%%%%%%%%%%%%%%%%%%%%%%%%%%%%%%%%%%%%%%%%%%%%%%%%
% APPENDIX
%%%%%%%%%%%%%%%%%%%%%%%%%%%%%%%%%%%%%%%%%%%%%%%%%%%%%%%%%%%%%%%%%%%%%%%%%%%%%%%
%%%%%%%%%%%%%%%%%%%%%%%%%%%%%%%%%%%%%%%%%%%%%%%%%%%%%%%%%%%%%%%%%%%%%%%%%%%%%%%
\clearpage
\newpage

\appendix

\section{Impact Statement}\label{app:impact_statement}
\paragraph{Limitation}
Our method relies on obtaining the hidden states from all model layers during the training and evaluation phases. This requirement means full internal access to the model's architecture and intermediate representations is essential for optimization. Consequently, the method is applicable and can be optimized only for open-source models, where such internal states are accessible. In contrast, for closed-source or proprietary models, where access to internal hidden states is typically restricted, our method cannot be deployed to improve performance, limiting its applicability in those scenarios.

\paragraph{Ethics Statement}
We affirm that this research adheres to the ethical standards in AI and machine learning. No personal, sensitive, or proprietary data was used in the experiments. All datasets employed are publicly available and widely used in the research community. Our code is publicly available, and all experiments are reproducible. We have carefully considered the potential societal impacts and are committed to applying the method responsibly, ensuring it has a positive and controllable impact on society.

\paragraph{Future Societal Consequences}
Our approach relies on comprehensive access to the internal states of models, which has profound implications for the future ecological landscape of AI development. As open-source models continue to evolve, this method helps drive the democratization of performance optimization, enabling researchers to improve models transparently and collaboratively, fostering innovation, and enhancing the accountability of AI systems. However, it cannot be applied to closed-source or proprietary models, highlighting the growing gap between open and closed AI ecosystems and potentially increasing dependence on a few dominant providers. Furthermore, since this approach is primarily used for improving judgment models, there is a potential risk of misuse, such as interfering with or manipulating evaluation results, reminding us to be mindful of possible negative impacts when promoting its application, ensuring responsible and controlled use.

\section{Related Work}

\subsection{Evaluation of Text Generation Tasks}
Over the past decades, automatic evaluation metrics, including statistical methods and embedding-based approaches, have played a crucial role in assessing the quality of generated text. The metrics BLEU~\citep{papineni-etal-2002-bleu} and ROUGE~\citep{lin-2004-rouge} have long been standard for evaluating generated text. BERTScore~\citep{bertscore} evaluates the similarity between two sentences by using a pre-trained BERT model to generate contextual embeddings for each token and calculating the sum of their cosine similarities. BARTscore~\citep{bartscore} evaluates the quality of generated text by calculating the weighted logarithmic probability of one text given another text. \citet{chung-etal-2023-increasing} found that these metrics cannot be accurately evaluated for semantically equivalent expressions but syntactically different. Therefore, trying to evaluate harder-generated text requires more efficient methods. GPTScore~\citep{gptscore} evaluates text quality through conditional generation probabilities, enabling customizable assessments using natural language instructions. However, these methods all require reference answers for a stable evaluation.

\subsection{LLM-as-a-Judge}
The advancement of LLMs has made LLM-as-a-judge possible, where an LLM-based evaluator judges the model responses based on user instructions and the criteria from different aspects of the prompt. 
% The LLM evaluator dynamically adjusts the evaluation criteria and reveals an interpretable judgment process with chain-of-thought reasonings \citep{xu2024largelanguagemodelsactive}. It is more effective than statistics-based and embedding-based evaluators while cheaper and more efficient than human-based evaluations.
The evaluation is flexible as the evaluation criteria can be adjusted dynamically in the prompt instead of retraining the evaluator from scratch. The judgment can be interpretable and detailed criticism as well as feedback are generated before the judge score. The LLM-based evaluator is more efficient and scalable than human experts at much less expense. It is more effective and applicable than statistical evaluation metrics like BLEU and embedding-based metrics like BERTScore.

% %% llm judge的意义和用处
% The LLM judges are applied in different scenarios across response evaluation, model finetuning, and data synthesis. In addition to the evaluation and comparison of model responses, LLM evaluators serve as reward models and provide verification as well as feedback during the model enhancement finetuning. When used in data synthesis, LLM evaluators assess and filter the synthesized data. They are also required during the synthesis of preference data to annotate the accepted and rejected responses. The pair-wise LLM judges can be used in preference data synthesis, while point-wise judges are required in other applications.

%% llm judge的分类，基于prompt的，基于finetune的。
Now there are three different categories of LLM judges: The Point-wise judge model \citep{Prometheus,LLM-eval,llmpersonalizedjudge} evaluates individual responses independently, scoring them based on predefined criteria; The pair-wise judge model\citep{li2024generative,prometheus2,jung2024trustescalatellmjudges} compares two responses directly, determining which one is superior based on a set of evaluation dimensions; The List-wise judge model\citep{aligninghumanjudgmentrole} ranks multiple responses in terms of quality, assigning a rank to each based on the overall evaluation.
Researchers explore effective approaches to apply LLMs as judges by prompt-based or finetuning-based methods in various scenarios. The prompt-based approaches aim to enhance judgment via step-by-step instructions as well as multi-turn optimizations. G-Eval \citep{g-eval} evaluates NLG output quality using Chain-of-Thought (CoT) and Form-Filling paradigms. Portia \citep{li-etal-2024-split} proposes to evaluate the responses with step-by-step reasoning guidelines.
Active-critic \citep{xu2024largelanguagemodelsactive} first generates evaluation criteria from data and iteratively refines them. However, the performance of prompt-based LLM evaluators is unsatisfactory \citep{gu2024surveyllmasajudge}.
The finetuning-based approaches further optimize the LLMs with the specialized dataset to adapt the LLMs to judgment tasks either with instruction tuning \citep{ye2024selfj,li2024generative} or preference tuning \citep{ye2024conj,hu-etal-2024-themis}. 
InstructScore~\citep{instructscore} combines human guidance with implicit knowledge from models like GPT-4, finetuned on the LLaMA model, to provide both scores and human-readable diagnostic reports. TIGERScore~\citep{TIGERScore} is finetuned on the LLaMA-2 model to generate detailed error analysis, helping users understand each identified error and its associated penalty score.
Prometheus \citep{Prometheus} and Prometheus 2 \citep{prometheus2} are finetuned for fine-grained evaluation, achieving new state-of-the-art performance on judgment benchmarks.
% Additionally, probability calibration methods that do not require fine-tuning can help align the judge model more closely with human evaluations. \citet{daynauth2024aligningmodelevaluationshuman} uses Bayesian statistics and statistical tests to identify and correct inherent biases in automatic evaluation metrics. ProbDiff~\citep{xia2024languagemodelsevaluateprobability} assesses which model is more proficient at handling a specific instruction by comparing the probability differences between two candidate LLMs on the same query. A larger probability difference indicates lower confidence in the generated output, suggesting relatively poorer performance. CRISPR~\citep{yang2024mitigatingbiasesinstructionfollowinglanguage} automatically identifies biased outputs, classifies neurons influencing them as biased neurons using interpretability methods, and eliminates detected biased neurons through structured pruning. But these methods still cannot be applied to point-wise evaluation tasks for a single-LLM.
% In this work, we focus on the single-LLM point-wise evaluator where no reference response is available. Our proposed \mname method enhances the judgment performance different from the prompt-based and finetuning-based method. 
In this work, we focus on the single-LLM point-wise evaluator without a reference response. Our proposed \mname method improves judgment performance by estimation with aggregated layer-wise logits, which is different from the idea of prompt-based and finetuning-based methods.

\section{Addition Experimental Results}
\begin{figure}[t]
    \centering
    \includegraphics[width=0.9\linewidth]{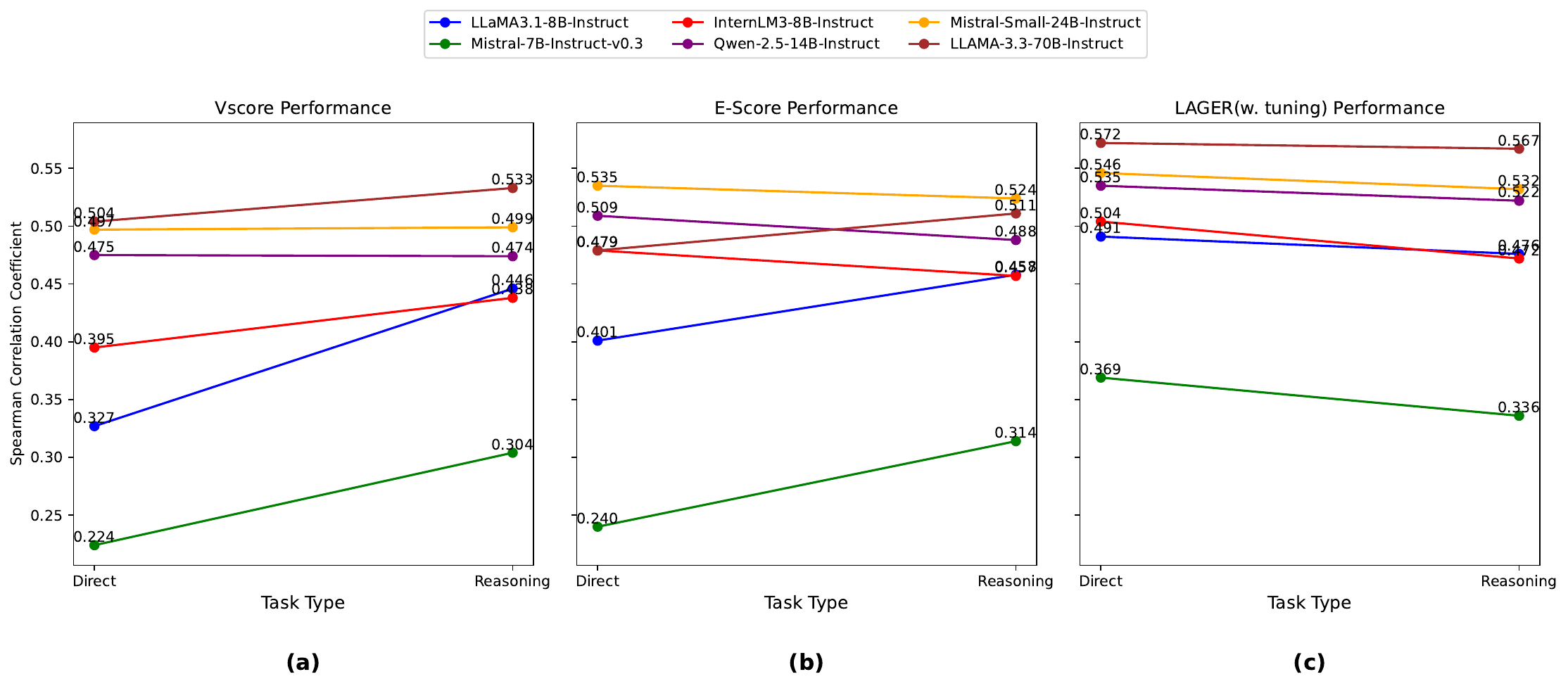}
    \caption{Average performance comparison of different scoring methods across models on direct and reasoning tasks over three benchmarks.}
    \label{fig:direct_vs_reasoning}
\end{figure}

\begin{figure}
    \centering
    \includegraphics[width=0.9\linewidth]{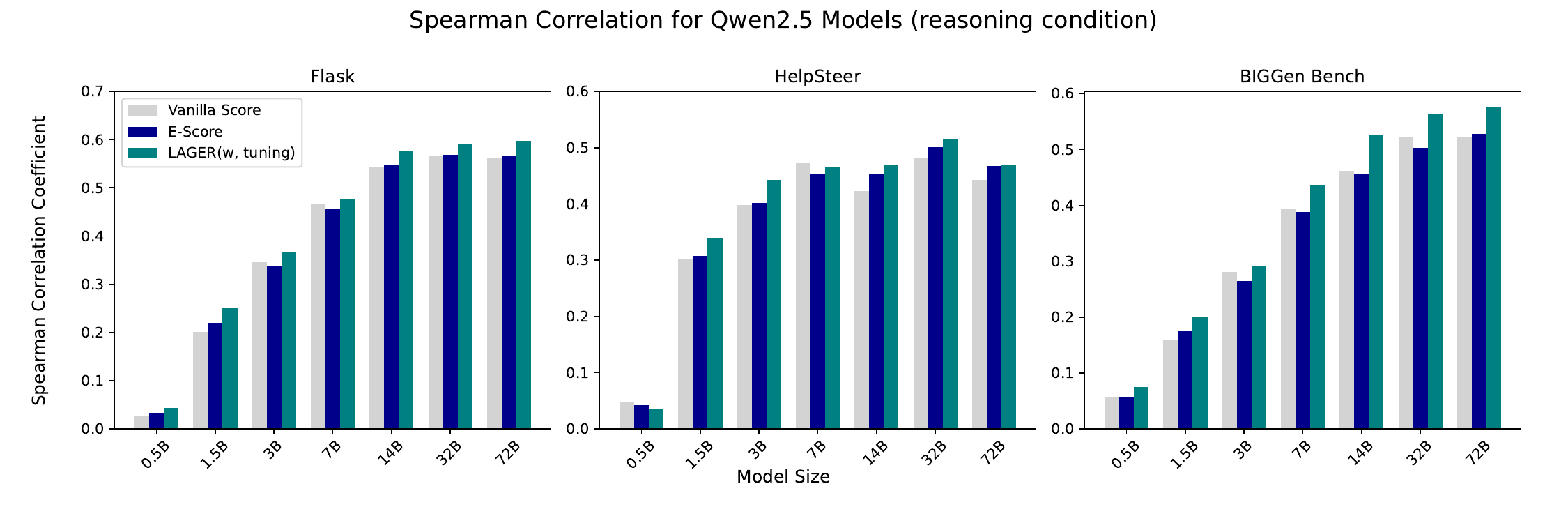}
    \caption{Comparison of Spearman Correlation Coefficients for Qwen2.5 Models Across FLASK, HelpSteer and BIGGen Bench Using Vanilla Score, E-Score, and \mname(w. tuning) (Reasoning Condition)}
    \label{fig:qwens-reasoning}
\end{figure}

In this section, Figure~\ref{fig:direct_vs_reasoning} shows the comparison of the performance of different scoring methods across models on direct and reasoning conditions. Figure~\ref{fig:qwens-reasoning} shows the comparison of spearman correlation coefficients for Qwen2.5 models across FLASK, HelpSteer and BIGGen Bench Using Vanilla Score, E-Score, and \mname(w. tuning) under the reasoning condition.  In Table~\ref{table:pearson results}, we present the Pearson correlation coefficients for all models, including \mname and other baselines, on the Flask, HelpSteer, and BiGGen Bench. 

\begin{table*}[t]

\centering
\caption{Pearson correlation coefficients of \mname and other baselines on Flask, HelpSteer, and BiGGen Benchmarks. These results are statistically significant, with p-values less than 1e-5.  \textit{These
results are statistically significant, with p-values less than 1e-5}.}
\resizebox{\textwidth}{!}{\begin{tabular}{llccccccc}
\toprule
\textbf{Model} &  & \begin{tabular}[c]{@{}c@{}}Flask\\ Direct\end{tabular} & \begin{tabular}[c]{@{}c@{}}Flask\\ Reasoning\end{tabular} & \begin{tabular}[c]{@{}c@{}}HelpSteer\\ direct\end{tabular} & \begin{tabular}[c]{@{}c@{}}HelpSteer\\ Reasoning\end{tabular} & \begin{tabular}[c]{@{}c@{}}BIGGen Bench\\ Direct\end{tabular} & \begin{tabular}[c]{@{}c@{}}BIGGen Bench\\ Reasoning\end{tabular} & Average \\ \hline
\multicolumn{9}{c}{\cellcolor[HTML]{DAE8FC}\textit{\textbf{Fine-tuned Models}}} \\ \hline
\textbf{TIGERscore-7B} &  & - & 0.227 & - & 0.173 & - & 0.116 & 0.172 \\
\textbf{Prometheus2-7B} &  & - & 0.453 & - & 0.502 & - &  &  \\ \hline
\multicolumn{9}{c}{\cellcolor[HTML]{DAE8FC}\textit{\textbf{Close-source Model via API}}} \\ \hline
\textbf{GPT-4o-mini} &  &  &  &  &  &  &  &  \\ 
Vscore &  & 0.563 & 0.581 & 0.484 & 0.529 & 0.576 & 0.538 & 0.545 \\
E-Score &  & 0.588 & 0.587 & 0.503 & 0.532 & 0.588 & 0.543 &  0.557\\ \hline
\multicolumn{9}{c}{\cellcolor[HTML]{DAE8FC}\textit{\textbf{Open-ource Models}}} \\ \hline
\textbf{LLaMA3.1-8B-Instruct} &  &  &  &  &  &  &  &  \\ 
GPTScore &  & 0.109 & - & 0.028 & - & -0.109 & - & 0.009 \\
Vscore &  & 0.393 & 0.486 & 0.376 & 0.517 & 0.329 & 0.441 & 0.424 \\
E-Score &  & 0.386 & 0.494 & 0.427 & \textbf{0.525} & 0.380 & 0.452 & 0.444 \\
\mname(w.o tuning) &  & 0.487 & \textbf{0.514} & 0.483 & \textbf{0.525} & 0.473 & 0.481 & 0.494 \\
\mname(w. tuning) &  & \textbf{0.493}& 0.512 & \textbf{0.486}& 0.524& \textbf{0.485}& \textbf{0.482}& \textbf{0.497} \\ \hline
\textbf{Mistral-7B-Instruct-v0.3} &  &  &  &  &  &  &  &  \\ 
GPTScore &  & 0.336 & - & 0.166 & - & 0.154 & - & 0.219 \\
Vscore &  & 0.292 & 0.305 & 0.302 & 0.355 & 0.194 & 0.321 & 0.295 \\
E-Score &  & 0.307 & 0.313 & 0.333 & 0.363 & 0.221 & 0.329 & 0.311 \\
\mname(w.o tuning) &  & \textbf{0.380} & \textbf{0.343} & 0.412 & \textbf{0.380} & 0.357 & \textbf{0.376} & \textbf{0.375} \\
\mname(w. tuning) &  & 0.378 & 0.337 & \textbf{0.413} & 0.377 & \textbf{0.361} & 0.375 & 0.374 \\ \hline
\textbf{InternLM3-8B-Instruct} &  &  &  &  &  &  &  &  \\ 
GPTScore &  & -0.104 & - & -0.047 & - & -0.128 & - & -0.093 \\
Vscore &  & 0.468 & 0.516 & 0.374 & 0.429 & 0.399 & 0.475 & 0.444 \\
E-Score &  & 0.544 & 0.532 & 0.446 & 0.441 & 0.477 & 0.489 & 0.488 \\
\mname(w.o tuning) &  & 0.475 & 0.517 & 0.417 & 0.441 & 0.371 & 0.471 & 0.449 \\
\mname(w. tuning) &  & \textbf{0.568} & \textbf{0.551} & \textbf{0.494}& \textbf{0.474}& \textbf{0.497}& \textbf{0.501}& \textbf{0.512} \\ \hline
\textbf{Qwen-2.5-14B-Instruct} &  &  &  &  &  &  &  &  \\ 
GPTScore &  & -0.026 & - & -0.065 & - & -0.108 & - & -0.066 \\
Vscore &  & 0.585 & 0.581 & 0.428 & 0.420 & 0.494 & 0.507 & 0.503 \\
E-Score &  & 0.612 & 0.594 & \textbf{0.448} & 0.436 & 0.517 & 0.515 & 0.520 \\
\mname(w.o tuning) &  & 0.612 & 0.600 & 0.430 & 0.436 & 0.496 & 0.503 & 0.513 \\
\mname(w. tuning) &  & \textbf{0.645}& \textbf{0.618} & 0.443& \textbf{0.457} & \textbf{0.576}& \textbf{0.551}& \textbf{0.548} \\ \hline
\textbf{Mistral-Small-24B-Instruct} &  &  &  &  &  &  &  &  \\ 
GPTScore &  & -0.020 & - & -0.044 & - & -0.088 & - & -0.051 \\
Vscore &  & 0.573 & 0.545 & 0.418 & 0.440 & 0.589 & 0.575 & 0.523 \\
E-Score &  & \textbf{0.608} & 0.561 & 0.442 & 0.463 & 0.624 & 0.589 & 0.548 \\
\mname(w.o tuning) &  & 0.603 & \textbf{0.577}& 0.448 & 0.477 & 0.628 & 0.603 & 0.556 \\
\mname(w. tuning) &  & \textbf{0.608} & 0.569& \textbf{0.448} & \textbf{0.477} & \textbf{0.634} & \textbf{0.604} & \textbf{0.557} \\ \hline
\textbf{LLAMA-3.3-70B-Instruct} &  &  &  &  &  &  &  &  \\ 
GPTScore &  & 0.022 & - & -0.033 & - & -0.121 & - & -0.044 \\
Vscore &  & 0.544 & 0.607 & 0.435 & 0.482 & 0.590 & 0.587 & 0.541 \\
E-Score &  & 0.554 & 0.610 & 0.444 & 0.485 & 0.597 & 0.588 & 0.546 \\
\mname(w.o tuning) &  & \textbf{0.627} & \textbf{0.634} & \textbf{0.506} & \textbf{0.515} & 0.644 & \textbf{0.624} & \textbf{0.592} \\
\mname(w. tuning) &  & 0.626& 0.630 & 0.503 & 0.512 & \textbf{0.645} & 0.620 & 0.589 \\ \bottomrule
\end{tabular}
}
\label{table:pearson results}
\end{table*}

\section{Addition Applications of LAGER}
\label{sec:app_lager}

\subsection{Emotional Understanding} \label{sec:EU}
LLMs have demonstrated strong capabilities to comprehend and interpret complex emotions and their meanings in social contexts. Therefore, beyond merely assessing the quality of responses, LLM can also be used as an emotion judge. 
In the widely used emotion understanding benchmark EQ-Bench \citep{eqbench}, EQ-Bench is a benchmark for language models designed to assess emotional intelligence. It uses a specific question format in which participants are required to read a conversation and then rate the intensity of one character's emotional response. Each question is explanatory and aims to assess the ability to predict the intensity of four different emotions. The dataset originally contained 117 questions, and after processing, we obtain 430 instances, each corresponding to one emotion per conversation. The emotional intensity annotated by humans ranges from 1 to 9, with higher scores indicating stronger emotions. The LLM is asked to predict the intensity of the emotional states of characters in a dialogue with a score ranging from 1 to 9. In this part, we evaluate \mname on EQ-Bench and report the consistency with human evaluations. 

As shown in Table \ref{table:sentiment}, \mname significantly outperforms the widely used vanilla score across all LLM backbones and consistency metrics, with an average improvement of 8.42 points without tuning and 9 points with tuning. This demonstrates that \mname is effective not only for evaluating the quality of responses but also in other LLM-as-a-Judge scenarios, such as emotional understanding. \mname is tuned on the HelpSteer validation set, which has a different scoring domain and range compared to EQ-Bench (ranging from 1-5 versus 1-9). This further highlights the robustness of \mname with tuning, \textbf{as it successfully transfers from one domain and score range to other domains with different judging prompts and score ranges}.

\begin{table}[ht]

 \caption{The Spearman and Pearson correlation coefficient measures the consistency between reference sentiment intensity and the scores from
    LLaMA3.1-8B-Instruct, Mistral-7B-Instruct-v0.3, Qwen-2.5-14B-Instruct, on the processed EQ-Bench.}
\centering
\fontsize{8}{10}\selectfont
\resizebox{0.6\linewidth}{!}{
\begin{tabular}{lccc}
\toprule
\multirow{2}{*}{\textbf{Model}} & \multirow{2}{*}{\textbf{Scoring Type}} & \multicolumn{2}{c}{\textbf{Processed-EQ-Bench}} \\ \cline{3-4} 
 &  & Pearson & Spearman \\ \hline
\multirow{4}{*}{Mistral-7B-Instruct-v0.3} & VScore & 0.56 & 0.596 \\
&E-Score & 0.574 & 0.618 \\
 & \mname(w.o. tuning) & 0.593 & 0.632 \\
 & \mname(w. tuning) & \textbf{0.598} & \textbf{0.635} \\\hline
\multirow{4}{*}{LLaMA-3.1-8B-Instruct} & VScore & 0.456 & 0.478 \\
 &E-Score & 0.608 & 0.652 \\
 & \mname(w.o. tuning) & 0.627 &  0.653\\ 
 & \mname(w. tuning) & \textbf{0.634} & \textbf{0.657} \\\hline
 \multirow{4}{*}{Qwen-2.5-14B-Instruct} & VScore & 0.706 & 0.707 \\
 &E-Score & 0.739 & 0.759 \\
 & \mname(w.o. tuning) & 0.745 &  0.758\\ 
 & \mname(w. tuning) & \textbf{0.756} & \textbf{0.763} \\
 \bottomrule
\end{tabular}}
\label{table:sentiment}
\end{table}

\subsection{Enhancing LLM in Knowing When They Don't Know} \label{sec:selfaware}

Knowing when they don't know, also known as self-knowledge \citep{yin-etal-2023-large}, is crucial for enhancing the reliability and trustworthiness of LLMs. In this section, we explore whether \mname can improve an LLM's self-knowledge. We evaluate the SelfAware dataset \citep{yin-etal-2023-large} using LLaMA3.1-8B-Instruct and Mistral-7B-Instruct-v0.3, which consists of 2337 answerable and 1032 unanswerable questions from five diverse categories. 
Specifically, given a subjective question in SelfAware, we ask an LLM to evaluate the answerability of this question on a scale of 1 to 5. A score of 1 indicates that the question is completely unanswerable based on the available knowledge, while a score of 5 indicates a highly accurate answer is possible. Following the original paper, we normalize the certainty score to a range of 0-1 and use a threshold of 0.75 as the boundary to classify LLM as either \textit{"know"} ($\geq$ 0.75) or \textit{"don't know"} ($<$ 0.75). 

\begin{wraptable}{r}{0.5\textwidth}
\centering
\fontsize{8}{10}\selectfont
\caption{Self-knowledge Comparison of \mname and baselines. The evaluation metric is the F1-Score. }
\resizebox{\linewidth}{!}{
\begin{tabular}{@{}lccccc@{}}
\toprule
\multirow{3}{*}{\textbf{Model}} & \multicolumn{5}{c}{\textbf{Classification Methods}} \\ \cmidrule(l){2-6} 
 & SimCSE & VScore & E-Score& \begin{tabular}[c]{@{}c@{}}\mname\\ (w.o tuning)\end{tabular} & \begin{tabular}[c]{@{}c@{}}\mname\\ (w. tuning)\end{tabular} \\ \midrule
 Mistral-7B-Instruct-v0.3 & 0.427 & 0.605 & 0.62 & 0.62 & \textbf{0.624} \\
LLaMA-3.1-8B-Instruct & 0.490 & 0.553 &0.574& \textbf{0.59} & 0.570 \\
 \bottomrule
\end{tabular}}
\label{table:certainty}
\end{wraptable}
Then, we can quantitatively calculate the F1 score to measure the model's level of self-knowledge.Table \ref{table:certainty} compares \mname with baseline methods, including the SimCSE baseline used in the original paper. The results show that \mname (w. tuning) significantly enhances the LLM's self-knowledge, achieving a +19.7\% and +1.9\% increase in F1 score compared to SimCSE, and VScore on LLaMA-3.1-8B-Instruct, respectively, and +11.1\%, and +4.8\% increases in F1 score compared to SimCSE and VScore on Mistral-7B-Instruct-v0.3, respectively. By providing a more reliable measure of LLM's self-knowledge, \mname enhances the LLM's ability to understand their limitations, shows performance comparable to the E-score.

\section{Experiment Details} \label{sec:app_details}

\subsection{Details About KDE for Score Distribution Visualization}
\label{sec:app_kde}
We adopt kernel density estimation (KDE) to visualize the distribution of model-generated scores across evaluation samples. KDE provides a smooth, non-parametric estimate of the underlying score distribution and is preferable to histograms for highlighting fine-grained density differences.

Given a set of samples \( \{x_1, x_2, \dots, x_n\} \), the estimated density at point \( x \) is computed as:
\begin{equation}
    \hat{f}_h(x) = \frac{1}{n h} \sum_{i=1}^{n} K\left( \frac{x - x_i}{h} \right),
\end{equation}
where \( K(\cdot) \) is the kernel function and \( h \) is the bandwidth. We use the Gaussian kernel:
\begin{equation}
    K(u) = \frac{1}{\sqrt{2\pi}} e^{-u^2/2}.
\end{equation}

All KDE plots are normalized to unit area for comparability. We employ kernel density estimation for visualization, with the bandwidth parameter selected using Silverman's rule or cross-validation when appropriate.
KDE is used in Figure~\ref{fig:estimated-kernel-density} to compare the score output distributions of \mname and baseline models.

\subsection{Details About the Benchmarks} \label{sec:app_bench}
% \begin{itemize}

\paragraph{FLASK} We utilize the complete test prompt set from FLASK. \citep{flask}, a fine-grained evaluation dataset that includes various conventional NLP datasets and instruction datasets. This dataset contains data on 2,001 entries, each consisting of a human feedback score and an evaluation score from GPT-4. We utilize 12 scoring rubrics: Conciseness, Metacognition, Insightfulness, Readability, Commonsense Understanding, Logical Robustness, Factuality, Comprehension, Completeness, Logical Efficiency, Logical Correctness, and Harmlessness.

\paragraph{HelpSteer} HelpSteer \citep{wang2023helpsteermultiattributehelpfulnessdataset} is an open-source Helpfulness Dataset. For each response in the dataset, it is evaluated based on five criteria: helpfulness, correctness, coherence, complexity, and verbosity. These evaluation scores are annotated by humans, ensuring that each instruction in the dataset is paired with responses of varying quality. We process these data, combining the prompt, response, and corresponding criteria into individual data samples. A total of 8.95k data points are generated, with the first 2k used for our evaluation.

\paragraph{BiGGen Bench}BiGGen Bench~\citep{prometheus2} is a comprehensive evaluation benchmark designed to assess the capabilities of large language models (LLMs) across a wide range of tasks. This benchmark focuses on free-form text generation and employs fine-grained, instance-specific evaluation criteria. BiGGen Bench evaluates nine distinct generation capabilities (e.g., instruction following, reasoning, tool usage, etc.) across 77 tasks, providing model
outputs and scores for 103 different language models. We utilize the human evaluation test set.

% \paragraph{UltraFeedback-Binarized} This is a pre-processed version of the UltraFeedback dataset. The original UltraFeedback dataset~\citep{cui2023ultrafeedback} consists of 64K prompts, where each prompt is accompanied by four model completions from a wide variety of open and proprietary models. GPT-4 is then used to assign a score to each completion, along with criteria like helpfulness and honesty. To create UltraFeedback-Binarized, the highest overall score is picked as the chosen completion, and one of the remaining 3 at random is the rejected one. The UltraFeedback dataset used in this work is sampled from 2K instances in UltraFeedback-Binarized where the score difference between the chosen and rejected responses is at least 2.

% \paragraph{HelpSteer-Prefer} We process the test split of HelpSteer and HelpSteer2~\citep{helpsteer2}, extracting a total of 849 preference pairs as the HelpSteer-Prefer datasets. The filtering method is as follows: First, we calculate the average of three metrics—helpfulness, correctness, and coherence—for each response in the dataset, using this average as a comprehensive quality indicator for the response (complexity and verbosity were not included as they are not quality evaluation metrics). For any prompt where the comprehensive quality scores of the corresponding responses are not entirely consistent, the response with the highest score is selected as the chosen response, while the one with the lowest score is designated as the rejected response. 
% \end{itemize}

\subsection{Details About the Baselines} 
\label{app:baselines}
Due to the limitations of the GPTScore method, we do not report its results under the reasoning condition. Since TIGERScore-7B and Prometheus2-7B cannot directly output a single score, we do not report their results under the direct condition.
% \begin{itemize}                 
\paragraph{TIGERScore} TIGERScore\citep{TIGERScore} is a trained metric designed for explainable, reference-free evaluation of text generation tasks. Unlike other automatic evaluation methods that only provide scores, TIGERScore uses natural language instructions to guide error analysis, identifying specific mistakes in the generated text. It is based on the LLaMA-2 model and trained on a carefully curated dataset, MetricInstruct, which includes 42K quadruples covering 6 text generation tasks and 23 datasets. TIGERScore is capable of generating detailed error analysis, including the error location, error type, error explanation, and revision suggestions, along with a penalty score for each error.
    
\paragraph{GPTScore} Assuming the text to be evaluated is $\boldsymbol{h}=\{h_1, h_2,\cdots,h_m\}$, The core of the GPTscore method lies in leveraging LLM to evaluate the quality of text by computing the average log probability $\frac{1}{m}\sum_{t=1}^{m}\log p(h_t|\boldsymbol{h_{<t}}, T(d, a, S); \theta)$ under a specific context $S$ and task instruction $T(d, a, S)$. The prompt template $T(\cdot)$ is composed of a task description and aspect definition.
    
\paragraph{Prometheus-2-7B} The process of constructing the Prometheus-2 model~\citep{prometheus2} involves the following steps: First, a fine-grained pairwise ranking dataset, PreferenceCollection, containing 1000 custom evaluation criteria, is created to support evaluation based on user-defined standards. Next, Mistral-7B is selected as the base model, and it is trained separately on the direct evaluation dataset FEEDBACK COLLECTION and the pairwise ranking dataset PreferenceCollection. Finally, by merging the weights of the models trained on these two formats, the final Prometheus-2 model is obtained. 

\subsection{Details of Evaluation Metrics} \label{sec:app_metrics}

% \begin{itemize}
\paragraph{\textsc{Pearson}} Pearson correlation coefficient indicates the strength and direction of the relationship between two variables, ranging from -1 to 1. For the two sets of data x,y, we can calculate their Pearson correlation coefficients $\rho_{XY}$ as follows:
\begin{equation}
    \rho_{XY}=\frac{Cov(X,Y)}{\sigma_{X}\sigma_{Y}} = r_p
\end{equation}
    % $$\rho_{XY}=\frac{Cov(X,Y)}{\sigma_{X}\sigma_{Y}}$$

\paragraph{\textsc{Spearman}} Spearman rank correlation coefficient is a non-parametric measure that evaluates the statistical dependence between the ranks of two variables. It determines how well the relationship between these variables can be described using a monotonic function. We can calculate Spearman $r_s$ according to the following formula:
\begin{equation}
     r_s = 1- \frac{6\sum_{i=1}^{n} d^2}{n(n^2-1)}
\end{equation}
% $$ r_s = 1- \frac{6\sum_{i=1}^{n} d^2}{n(n^2-1)}$$
where $d_i$ represents the rank difference between the two variables for each observation (pairwise difference), and $n$ denotes the number of fraction pairs.
% \end{itemize}

Since the Spearman correlation coefficient effectively measures nonlinear monotonic relationships, is robust to outliers, and does not require data to follow a normal distribution, it is particularly suitable for handling ordinal data. Therefore, this work primarily presents results based on the Spearman correlation coefficient.

% \subsection{The Model's Vselect Results}
% Table~\ref{tab:vselect-res} presents the results of InternLM3-8B-Instruct, LLaMA3.1-8B-Instruct, and Mistral-7B-Instruct-v0.3 on HelpSteer-Prefer and UltraFeedback VSelect. The results indicate that InternLM3-8B-Instruct achieves the best performance across both datasets, while Mistral-7B-Instruct-v0.3 performs the worst.
% \begin{table}[ht]
% \caption{The model's Vselect results on Helpsteer-prefer and UltraFeedback.}
% \vskip 0.15in
% \resizebox{\linewidth}{!}{
% \begin{tabular}{@{}lcc@{}}
% \toprule
% \textbf{Model}                    & \textbf{Helpsteer-prefer} & \textbf{UltraFeedback} \\ \midrule
% InternLM3-8B-Instruct    & 0.736            & 0.846         \\
% LLaMA3.1-8B-Instruct     & 0.694            & 0.786         \\ 
% Mistral-7B-Instruct-v0.3 & 0.697            & 0.744         \\ \bottomrule
% \end{tabular}}
% \label{tab:vselect-res}
% \end{table}
\subsection{Details of layer-wise weight training settings} \label{sec:app_training_layer_weights}

We randomly select 1,000 samples from the HelpSteer dataset as a held-out validation set, ensuring that it does not overlap with the test set, to tune the layer-wise weights of \mname. The training is performed using the Adam optimizer with an initial learning rate of 0.01, and a batch size of 4. A random seed of 42 is set to ensure the reproducibility of the experiment. We also apply the ReduceLROnPlateau learning rate scheduler with a decay factor of 0.5, a patience value of 1, and a minimum learning rate specified by \texttt{min\_lr}. Since some models in the Qwen2.5 family may not converge with just one epoch, we set the number of training epochs to 2 for all models in the Qwen2.5 series. For other backbone models, we set the number of training epochs to 1. All experiments are implemented using PyTorch.

\section{Details of SFT}
\label{sec:app_sft}
For \mname, E-Score and VScore, we establish seven dimensions (including \textbf{answer accuracy, logical consistency, relevance, fluency and clarity, length appropriateness, diversity, and instruction difficulty}) and utilize the average score of each dimension, evaluated using the LLaMA3.1-8B-Instruct model, for data selection. The specific definitions and scoring standards for these seven dimensions can be found in Table~\ref{table:sft_filtering_dimensions}. SuperFiltering is an explicitly designed instruction data filtering method. We strictly follow the original paper's methodology, except for using LLaMA3.1-8B-Instruct as the backbone model for fair comparison. 
During the SFT training of the Llama-3-8B model, we use 4 L40 GPUs, and the total batch size during the training phase is 64. The specific training parameters and settings are detailed in Table~ \ref{table:sft-setting}. 
\begin{table}[t]
\caption{Parameters and settings used for instruction fine-tuning of the Llama-3-8B model.}
\centering
\vskip 0.15in
\rowcolors{2}{white}{gray!10} % 第2行开始，交替使用白色和浅灰色
\begin{tabular}{>{\raggedright}p{5cm}>{\raggedright\arraybackslash}p{3cm}}
\toprule
\textbf{Training Parameters} & \textbf{Setting} \\ \midrule
stage & sft \\ 
finetuning\_type & full \\ 
template & alpaca \\ 
flash\_attn & fa2 \\ 
cutoff\_len & 2048 \\ 
learning\_rate & 2e-5 \\ 
num\_train\_epochs & 3 \\ 
per\_device\_train\_batch\_size & 4 \\ 
gradient\_accumulation\_steps & 4 \\ 
lr\_scheduler\_type & cosine \\ 
warmup\_ratio & 0.03 \\ 
packing & FALSE \\ 
bf16 & TRUE \\ 
tf32 & TRUE \\ 
optim & adamw\_torch \\ 
include\_num\_input\_tokens\_seen & TRUE \\ 
seed & 42 \\ \bottomrule
\end{tabular}
\label{table:sft-setting}
\end{table}

\section{Prompt Template}
\label{sec:app_prompt_template}
\begin{figure}[ht]
    \centering
    \includegraphics[width=1\linewidth]{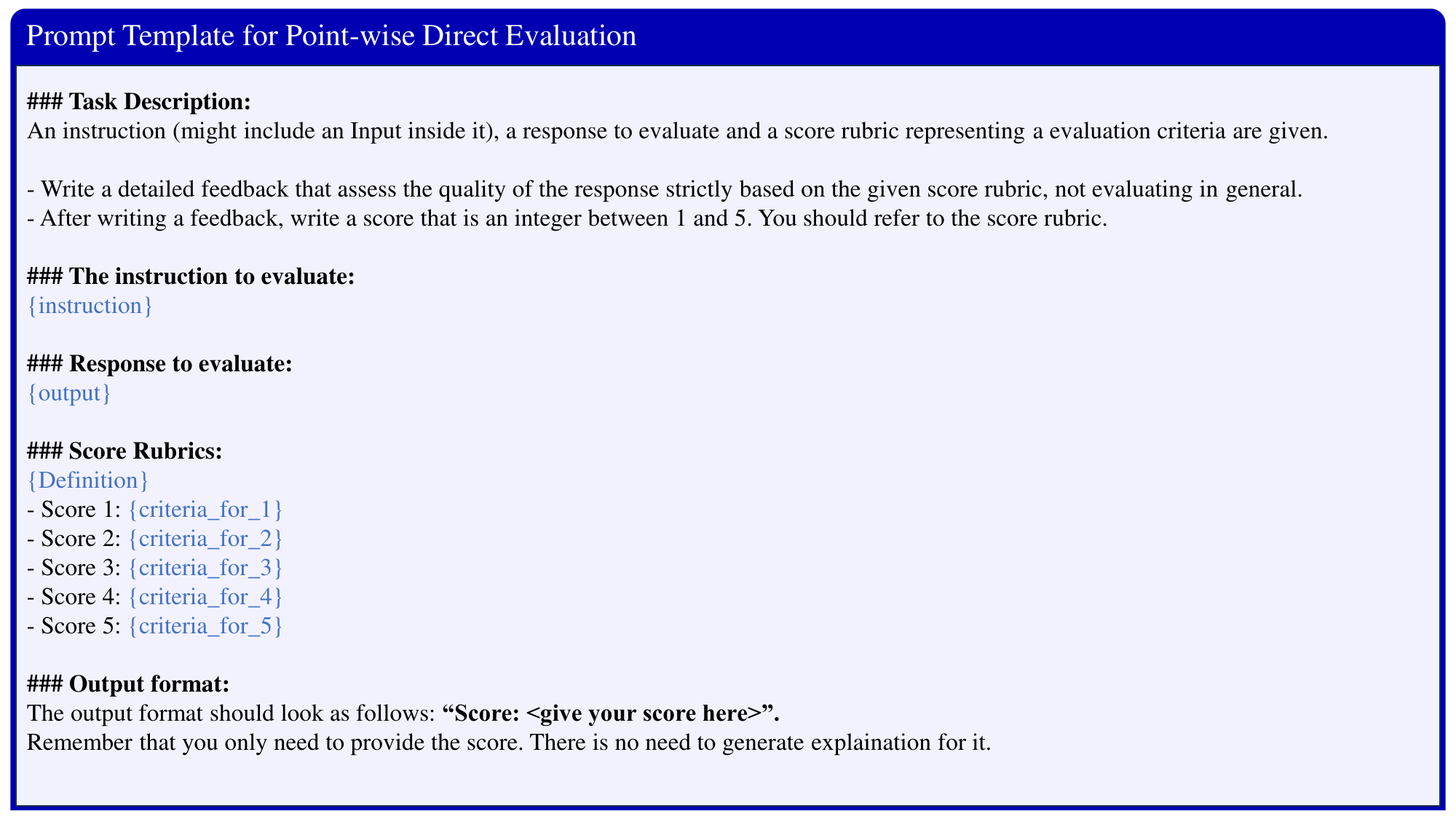}
    \caption{The prompt template for direct evaluation in a \textbf{point-wise} scenario that \textbf{outputs only a single score}.}
    \label{fig:point-wise-direct-template}
\end{figure}

\begin{figure}[ht]
    \centering
    \includegraphics[width=1\linewidth]{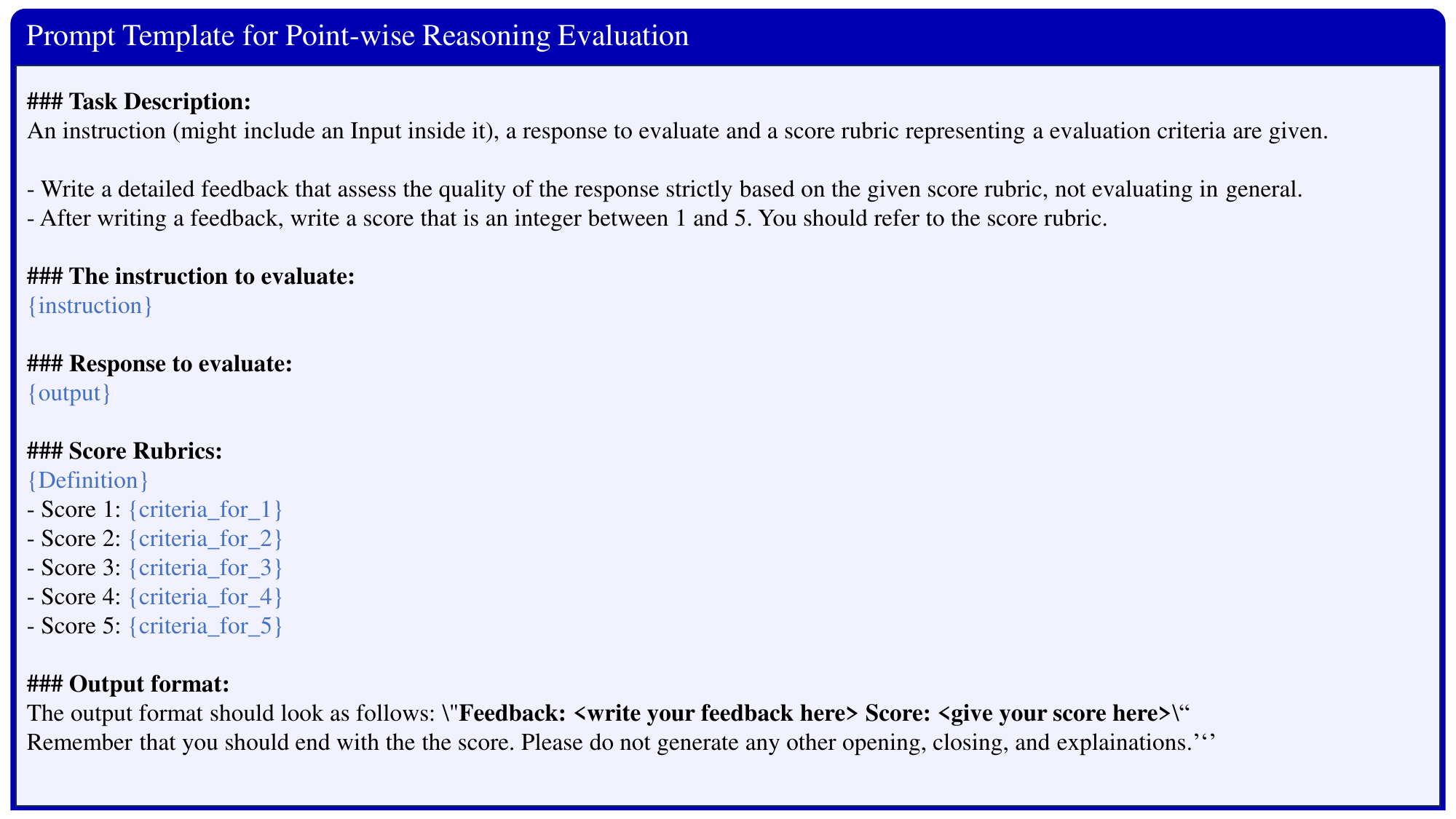}
    \caption{The prompt template for reasoning evaluation in a \textbf{point-wise} scenario, \textbf{including feedback(reasoning)}.}
    \label{fig:feedback-teplate}
\end{figure}

In this section, we present all the prompt templates used in this paper, with the blue text indicating the parts where corresponding content should be inserted. For the direct evaluation scenario,  Figure~\ref{fig:point-wise-direct-template} is the prompt template for direct evaluation in a point-wise
scenario that outputs only a single score. Figure~\ref{fig:feedback-teplate} is the prompt template for reasoning evaluation in a point-wise
scenario, including feedback(reasoning).

\begin{figure}[ht]
    \centering
    \includegraphics[width=1\linewidth]{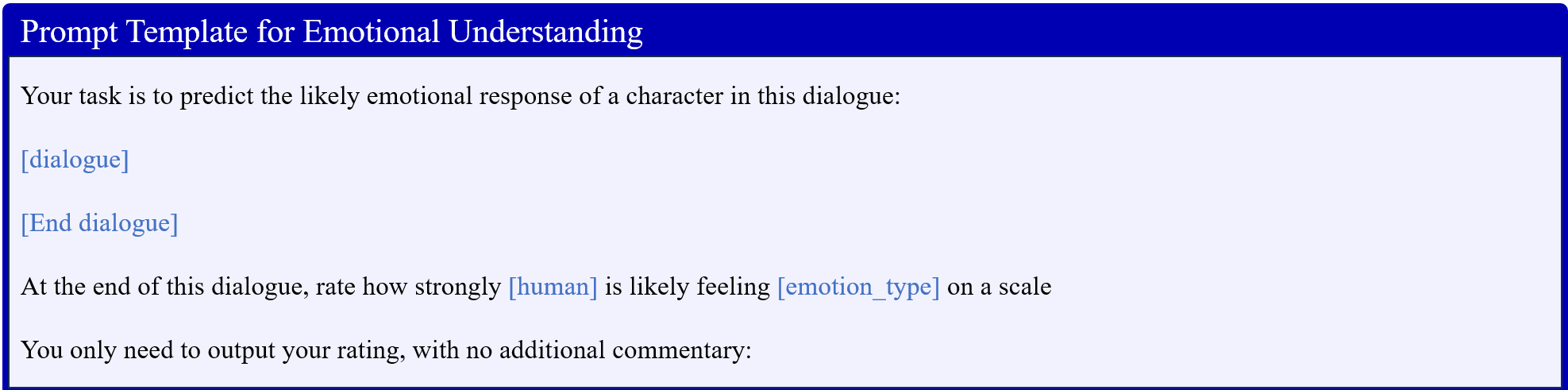}
    \caption{The prompt template for \textbf{Emotional Understanding} experiment.}
    \label{fig:eq-bench-template}
\end{figure}

% \begin{figure}[ht]
%     \centering
%     \includegraphics[width=1\linewidth]{pair.png}
%     \caption{The prompt template used for the \textbf{Pair} condition in the \textbf{pair-wise} datasets in the main experiment.}
%     \label{fig:pair-wise-Pair}
% \end{figure}

% \begin{figure}[ht]
%     \centering
%     \includegraphics[width=0.9\linewidth]{Vselect-direct.png}
%     \caption{The prompt template used for the \textbf{Vselect}  in the \textbf{pair-wise} datasets in the main experiment.}
%     \label{fig:vselect}
% \end{figure}

% \begin{figure}[ht]
%     \centering
%     \includegraphics[width=0.9\linewidth]{pair-wise-direct.png}
%     \caption{The prompt template used for the \textbf{Direct} condition in the \textbf{pair-wise} datasets in the main experiment.}
%     \label{fig:pair-wise-Single}
% \end{figure}

% \begin{figure}[ht]
%     \centering
%     \includegraphics[width=0.9\linewidth]{pair-wise-reasoning.png}
%     \caption{The prompt template used for the \textbf{Reasoning} condition in the \textbf{pair-wise} datasets in the main experiment.}
%     \label{fig:pair-wise-Reasoning}
% \end{figure}

\begin{figure}[ht]
    \centering
    \includegraphics[width=1\linewidth]{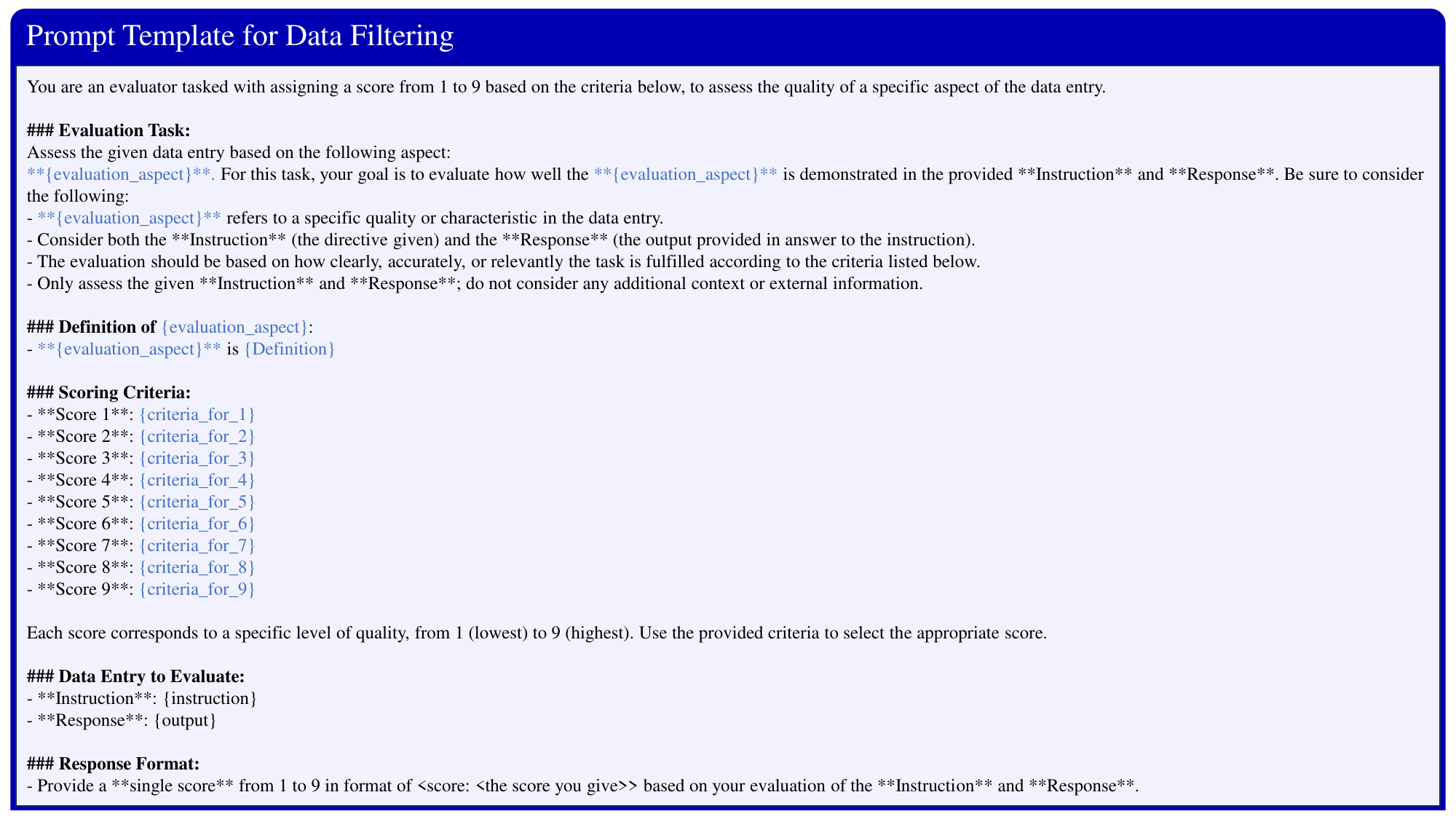}
    \caption{The prompt template used for data filtering in \textbf{SFT} task.}
    \label{fig:SFT}
\end{figure}

\begin{figure}[ht]
    \centering
    \includegraphics[width=1\linewidth]{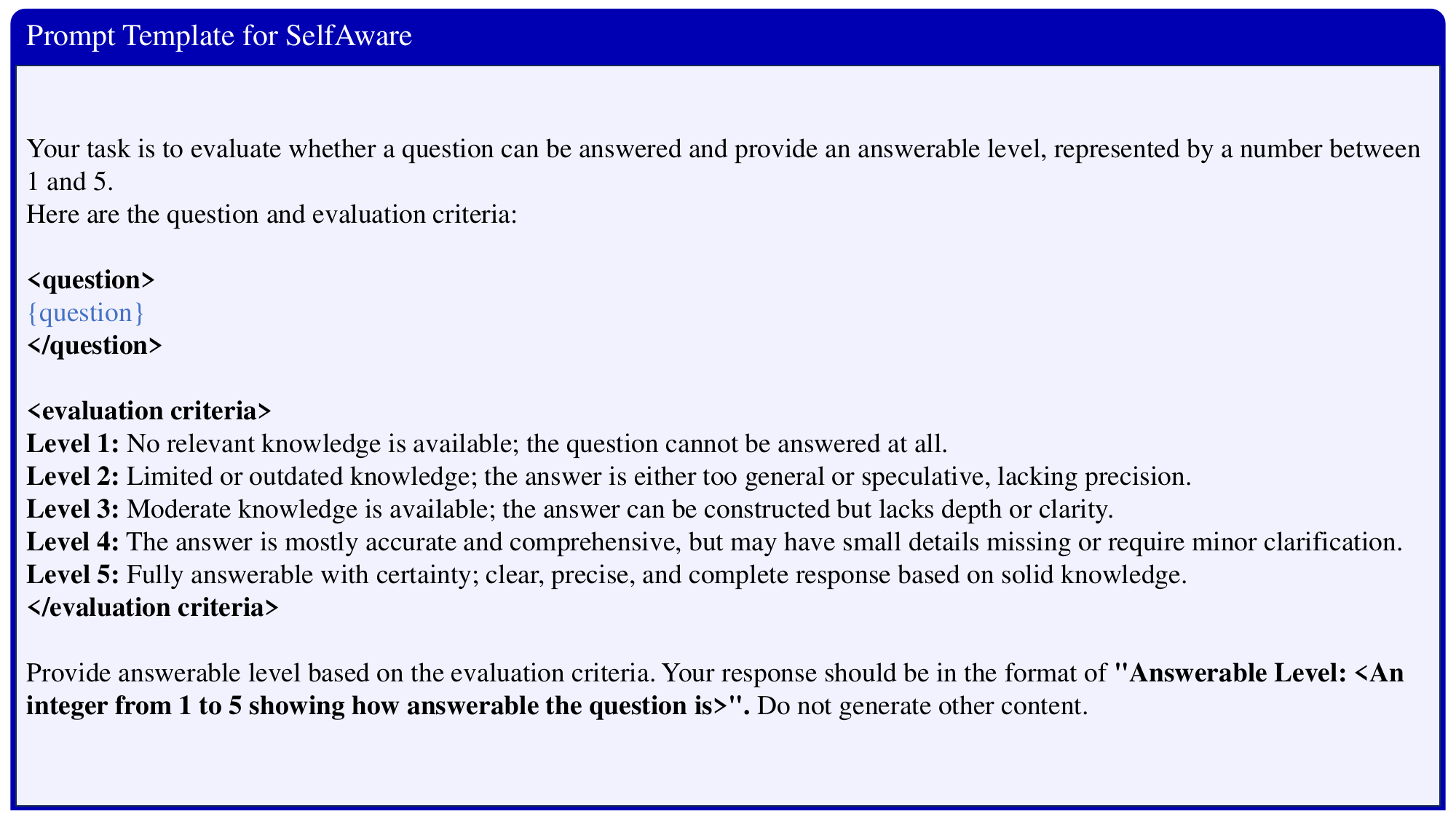}
    \caption{The prompt template used in the \textbf{SelfAware} dataset in section 7.2.}
    \label{fig:selfAware}
\end{figure}
\FloatBarrier

\begin{longtblr}[
  caption = {Definitions of different dimensions and their specific scoring standards, with each dimension scored on a scale ranging from 1 to 9, each score corresponds to a specific standard.},
]{
  width = \linewidth,
  colspec = {Q[150]Q[812]},
  row{4} = {t},
  column{1} = {c},
  cell{2}{1} = {r=9}{},
  cell{12}{1} = {r=9}{},
  cell{22}{1} = {r=9}{},
  cell{32}{1} = {r=9}{},
  cell{42}{1} = {r=9}{},
  cell{52}{1} = {r=9}{},
  cell{62}{1} = {r=9}{},
  vlines,
  hline{1,71} = {-}{0.12em},
  hline{2,11-12,21-22,31-32,41-42,51-52,61-62} = {-}{},
  hline{3-10,13-20,23-30,33-40,43-50,53-60,63-70} = {2}{},
}
\textbf{Dimension-1 \&~Definition} & \textbf{\textbf{~Answer Accuracy:~}}Evaluate whether the response accurately addresses the instruction and completely fulfills the task.\\
\textbf{\textbf{Scoring Standards}} & 1: Completely incorrect, irrelevant to the instruction.\\
 & 2: Partially correct, major omissions or errors in fulfilling the instruction.\\
 & 3: Contains significant errors, unable to fully address the core task.\\
 & 4: Partially correct, missing key details or addressing the wrong aspect of the instruction.\\
 & 5: Mostly accurate, but contains some errors or omissions.\\
 & 6: Mostly correct, though missing small details or has minor inaccuracies.\\
 & 7: Largely accurate and complete, but may lack small details or have minimal errors.\\
 & 8: Fully accurate, completely addresses the instruction, minimal flaws.\\
 & 9: Perfectly accurate, fully aligns with the instruction, no omissions.\\
\textbf{Dimension-2 \&~Definition} & \textbf{Logical Consistency}: Assess whether the response maintains logical consistency, following the reasoning of the instruction without contradictions.\\
\textbf{\textbf{\textbf{\textbf{Scoring Standards}}}} & 1: Completely incoherent, self-contradictory, does not follow the instruction.\\
 & 2: Major logical errors or contradictions, does not follow the instruction.\\
 & 3: Logic is unclear, unable to fully support the instruction's requirements.\\
 & 4: Contains some logical inconsistencies, but overall understandable, with minor deviation from the instruction.\\
 & 5: Mostly consistent, but has slight logical flaws or areas that could be clearer.\\
 & 6: Largely logical and clear, but small inconsistencies or areas for improvement exist.\\
 & 7: Logical and coherent, only minimal inconsistencies remain.\\
 & 8: Completely logical, reasoning is sound and aligns perfectly with the instruction.\\
 & 9: Perfectly logical, tightly reasoned, flawless adherence to the instruction's requirements.\\
\textbf{Dimension-3 \&~Definition} & \textbf{Relevance}: Evaluate whether the response is relevant to the instruction, directly addressing the key aspects of the task.\\
\textbf{\textbf{\textbf{\textbf{Scoring Standards}}}} & 1: Completely irrelevant, deviates from the instruction's theme.\\
 & 2: Minimal relevance, strays far from the core task.\\
 & 3: Poor relevance, does not adequately address the main parts of the instruction.\\
 & 4: Somewhat relevant, but deviates from key details or task elements of the instruction.\\
 & 5: Mostly relevant, but lacking some precision or key aspects of the instruction.\\
 & 6: Mostly relevant, covers the main aspects of the instruction, but can be improved for accuracy or detail.\\
 & 7: Directly relevant, clearly and adequately responds to the instruction.\\
 & 8: Highly relevant, fully addresses the core aspects of the instruction with precision.\\
 & 9: Perfectly relevant, fully and comprehensively addresses all aspects of the instruction.\\
\textbf{Dimension-4 \&~Definition} & \textbf{Fluency Clarity}: Evaluate the fluency and clarity of the response, ensuring it follows the expression requirements of the instruction and is easy to understand.\\
\textbf{\textbf{\textbf{\textbf{Scoring Standards}}}} & 1: Completely unclear, frequent grammar errors, hard to understand.\\
 & 2: Very unclear, poor structure, difficult to follow.\\
 & 3: Unclear, sentence structure issues, parts are hard to understand.\\
 & 4: Somewhat unclear, lacks fluency, but overall understandable.\\
 & 5: Clear, but could be improved in flow or conciseness.\\
 & 6: Clear and fluent, but slight improvements in clarity or conciseness could be made.\\
 & 7: Fluent, easy to understand, clear expression.\\
 & 8: Very clear and fluent, perfect adherence to the instruction's expression requirements.\\
 & 9: Perfectly fluent, natural, and clear, fully aligned with the instruction's expectations.\\
\textbf{Dimension-5 \&~Definition} & \textbf{Length Appropriateness}: Evaluate whether the response length is appropriate, providing sufficient information without being too brief or too long according to the instruction.\\
\textbf{\textbf{\textbf{\textbf{Scoring Standards}}}} & 1: Extremely short or overly long, irrelevant to the instruction's expected length.\\
 & 2: Too short or too long, fails to meet the instruction's requirements.\\
 & 3: Too brief, lacks key details, or too lengthy with unnecessary information.\\
 & 4: Slightly short or slightly long, lacks important details or includes redundancy.\\
 & 5: Reasonable length, but could be more concise or include more details as per the instruction.\\
 & 6: Length is appropriate, but could be refined by removing unnecessary parts or adding small details.\\
 & 7: Length is well-suited, adequately covers the instruction's main points without redundancy.\\
 & 8: Perfect length, concise yet comprehensive, fully meets the instruction's requirements.\\
 & 9: Optimal length, precisely conveys all required details, no redundancy, aligns perfectly with the instruction.\\
\textbf{Dimension-6 \&~Definition} & \textbf{Diversity}: Evaluate whether the response shows diversity in language, structure, and viewpoints, and avoids repetition or overly uniform expressions as required by the instruction.\\
\textbf{\textbf{\textbf{\textbf{Scoring Standards}}}} & 1: Extremely repetitive, no variation, fails to meet the instruction's diversity requirements.\\
 & 2: Lacks diversity, repetitive or formulaic expression, no innovation.\\
 & 3: Some diversity, but much of the content is repetitive or uniform.\\
 & 4: Some variation, but significant repetition or lack of diverse viewpoints.\\
 & 5: Some diversity, but certain sections are somewhat repetitive or conventional.\\
 & 6: Largely diverse, offers different perspectives or expressions, but with minor repetition.\\
 & 7: Strong diversity, rich in varied language and structure, aligns with the instruction's expectations.\\
 & 8: Very diverse, with significant creativity and variety in language and perspective.\\
 & 9: Extremely creative, highly diverse, fully meets the instruction's innovation and diversity requirements.\\
\textbf{Dimension-7 \&~Definition} & \textbf{Instruction Difficulty}: Assess the complexity of the instruction, considering whether it requires deep reasoning, multiple steps, or specialized knowledge, and how well the response reflects the difficulty of the task.\\
\textbf{\textbf{\textbf{\textbf{Scoring Standards}}}} & 1: Very simple instruction, requires only basic information with no reasoning.\\
 & 2: Simple task, requires basic common knowledge or simple answers.\\
 & 3: Slightly more complex, requiring some background knowledge or understanding of specific content.\\
 & 4: Moderately complex, involving some reasoning or moderately difficult tasks.\\
 & 5: Instruction requires multiple steps or specialized knowledge across domains.\\
 & 6: Complex instruction requiring advanced reasoning or a wide range of knowledge.\\
 & 7: Highly complex, requires deep reasoning or tasks that span multiple domains.\\
 & 8: Very complex, involves multi-layered reasoning or highly specialized knowledge.\\
 & 9: Extremely complex, requiring professional-level knowledge or intricate reasoning to fulfill.
 \label{table:sft_filtering_dimensions}
\end{longtblr}
\end{document}